\documentclass[lettersize,journal]{IEEEtran}
\usepackage{amsmath,amsfonts}
\usepackage{algorithmic}
\usepackage{algorithm}
\usepackage{array}
\usepackage[caption=false,font=normalsize,labelfont=sf,textfont=sf]{subfig}
\usepackage{textcomp}
\usepackage{stfloats}
\usepackage{url}
\usepackage{verbatim}
\usepackage{graphicx}
\usepackage{cite}

\usepackage{amssymb}
\usepackage{booktabs}

\usepackage{enumitem}
\usepackage{times}
\usepackage{epsfig}
\usepackage{amsfonts}
\usepackage{bm}
\usepackage{bbm}
\usepackage{multirow}
\usepackage{makecell}
\usepackage{ulem}
\usepackage{bbding}
\usepackage[table]{xcolor}
\definecolor{Klein_Blue}{rgb}{0.0, 0.129, 0.6}
\usepackage[pagebackref,breaklinks,colorlinks,bookmarks=false]{hyperref}
\hypersetup{
    colorlinks=true,
    linkcolor=magenta,
    urlcolor=magenta,
    citecolor=Klein_Blue
    }

\DeclareMathOperator*{\argmax}{arg\,max}

\newcommand{\ignore}[1]{}
\definecolor{cGreen}{RGB}{100,180,100}
\definecolor{cRed}{RGB}{220,50,0}

\begin{document}

\title{Unified Sequence-to-Sequence Learning for \\
Single- and Multi-Modal Visual Object Tracking}

\author{Xin~Chen,
        Ben~Kang,
        Jiawen~Zhu,
        Dong~Wang,
        Houwen~Peng,
        and~Huchuan~Lu
        \IEEEcompsocitemizethanks{
     \IEEEcompsocthanksitem {Xin Chen, Ben Kang, Jiawen Zhu, Dong Wang, and Huchuan Lu  are with the Dalian University of Technology. Houwen Peng is with the Microsoft Research.} 
     \IEEEcompsocthanksitem {Corresponding author: Dong Wang, wdice@dlut.edu.cn}
     }
        }
        
%\author{IEEE Publication Technology,~\IEEEmembership{Staff,~IEEE,}
        % <-this % stops a space
%\thanks{This paper was produced by the IEEE Publication Technology Group. They are in Piscataway, NJ.}% <-this % stops a space
%\thanks{Manuscript received April 19, 2021; revised August 16, 2021.}}



\maketitle

\begin{abstract}
In this paper, we introduce a new sequence-to-sequence learning framework for RGB-based and multi-modal object tracking. First, we present SeqTrack for RGB-based tracking. It casts visual tracking as a sequence generation task, forecasting object bounding boxes in an autoregressive manner. This differs from previous trackers, which depend on the design of intricate head networks, such as classification and regression heads. SeqTrack employs a basic encoder-decoder transformer architecture. The encoder utilizes a bidirectional transformer for feature extraction, while the decoder generates bounding box sequences autoregressively using a causal transformer. The loss function is a plain cross-entropy.
Second, we introduce SeqTrackv2, a unified sequence-to-sequence framework for multi-modal tracking tasks. Expanding upon SeqTrack, SeqTrackv2 integrates a unified interface for auxiliary modalities and a set of task-prompt tokens to specify the task. This enables it to manage multi-modal tracking tasks using a unified model and parameter set. This sequence learning paradigm not only simplifies the tracking framework, but also showcases superior performance across 14 challenging benchmarks spanning five single- and multi-modal tracking tasks. 
The code and models are available at  \href{https://github.com/chenxin-dlut/SeqTrackv2}{https://github.com/chenxin-dlut/SeqTrackv2}.
\end{abstract}


\section{Introduction}
\begin{figure}[!t]
\begin{center}
\includegraphics[width=1\linewidth]{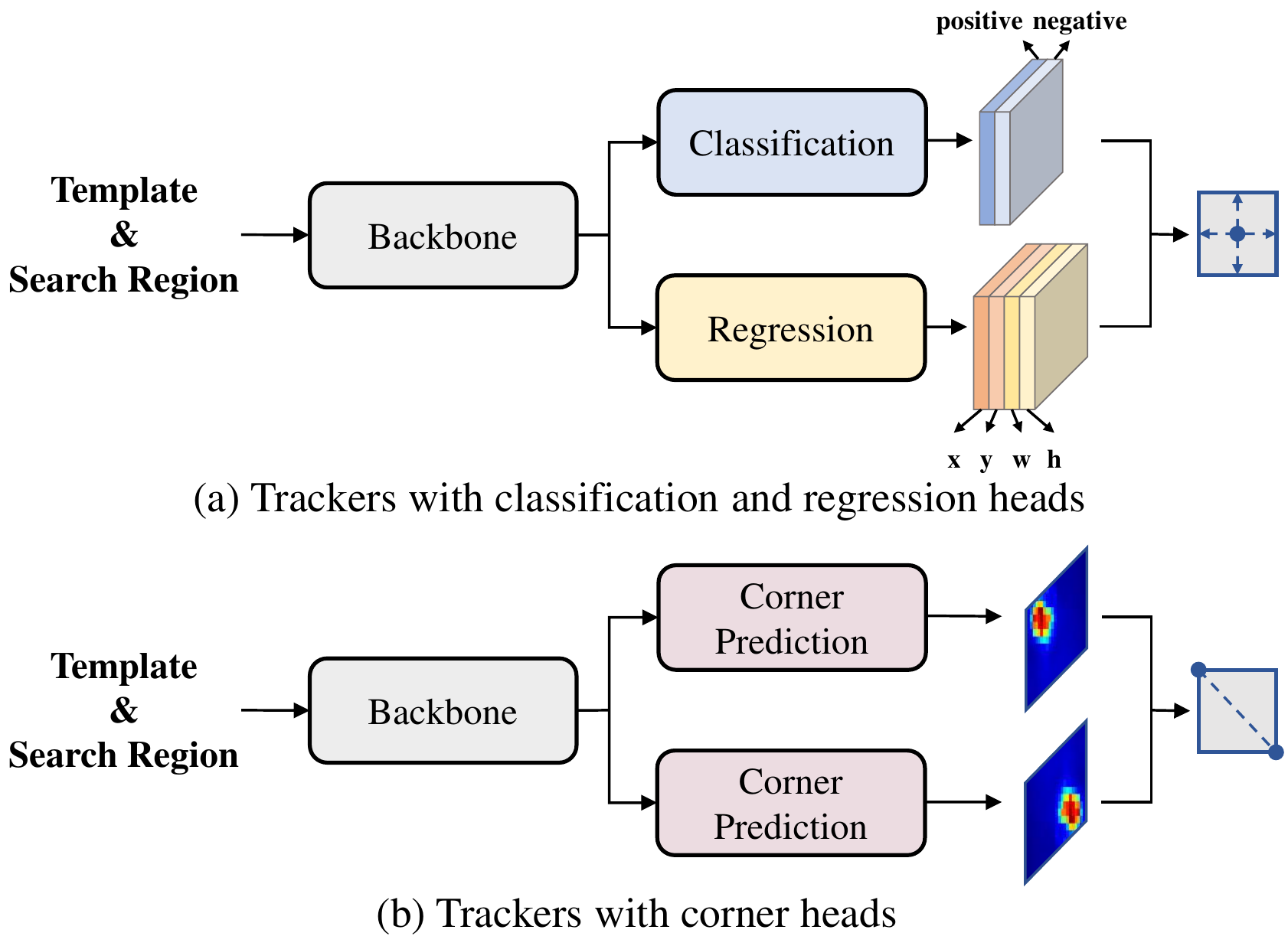}
\end{center}
\begin{center}
\includegraphics[width=0.88\linewidth]{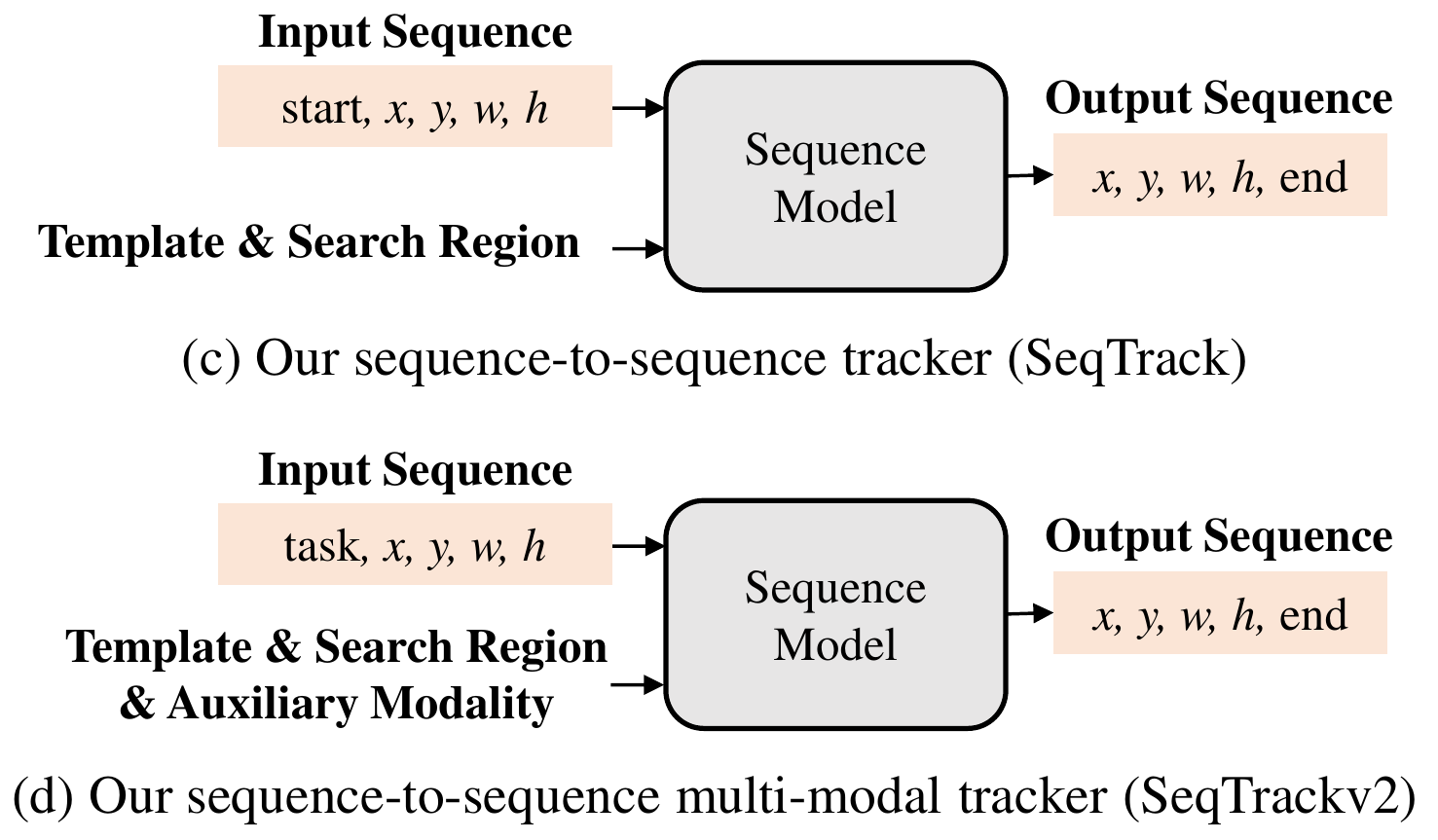}
\end{center}
    \vspace{-3mm}
   \caption{
   Comparison of tracking frameworks. (a) The framework with object classification head and bounding box regression head. (b) The framework with corner prediction heads. (c) Sequence-to-sequence tracking framework without complicated head networks. (d) Sequence-to-sequence framework for multi-modal tracking.} 
       \vspace{-5mm}
\label{fig:pipeline}
\end{figure}

\IEEEPARstart{V}isual object tracking is a foundational task in computer vision. Its objective is to estimate the position of an arbitrary target in  a video sequence, relying on its location in the initial frame. Existing tracking approaches commonly employ a divide-and-conquer strategy, wherein the tracking problem is decomposed into multiple subtasks, such as object scale estimation and center point localization.
Each subtask is addressed by a dedicated head network. For example, SiamRPN \cite{SiameseRPN} and its subsequent works \cite{DiMP, Ocean, transt, TMT, ostrack} utilize classification heads for object localization and regression heads for scale estimation, as illustrated in Fig.\ref{fig:pipeline}(a). STARK~\cite{Stark} and transformer-based trackers \cite{CSWinTT, simtrack, AiATrack} employ corner head networks to predict the bounding box corners of target objects, as depicted in Fig.\ref{fig:pipeline}(b).

Such a divide-and-conquer strategy has demonstrated superior performance on tracking benchmarks, becoming the prevalent design in current models. However, two deficiencies remain. First, each subtask necessitates a customized head network, leading to a complicated tracking framework. Second, each head network requires one or more learning loss functions, such as cross-entropy loss \cite{SiameseRPN, transt}, $\ell_1$ loss \cite{SiameseRPN, transt, Stark, ostrack}, and generalized IoU loss~\cite{transt, Stark, ostrack}, making the training challenging due to additional hyperparameters.

To address these issues, we propose a new Sequence-to-sequence Tracking (SeqTrack) framework in this paper, as illustrated in Fig.~\ref{fig:pipeline}(c). By modeling tracking as a sequence generation task, SeqTrack eliminates the need for complicated head networks and redundant loss functions. The underlying intuition is that if the model knows where the target object is, we could simply teach it how to read the bounding box out, rather than explicitly performing additional classification and regression via a divide-and-conquer strategy. 

To this end, we convert the four values of a bounding box into a sequence of discrete tokens and train the model to generate this sequence token-by-token. We utilize a simple encoder-decoder transformer to model this generation. The encoder extracts visual features from video frames, while the decoder generates the sequence of bounding box values using the extracted features. The generation process operates in an autoregressive manner, where the model generates a token based on previously observed ones.
At each step, a newly generated token value is fed back into the model to produce the next one. We apply a causal mask on the self-attention modules in the decoder to prevent tokens from attending to subsequent tokens. This causal masking mechanism ensures that the generation of the token at position $i$ only depends on its preceding tokens at positions before $i$. The visual features are integrated into the decoder through cross-attention layers \cite{2017Attention}. The generation concludes when it outputs four token values representing the bounding box. The resulting output sequence is directly used as the final tracking result.

Given the concise and versatile nature of the proposed sequence-to-sequence tracking framework, it is easily extended. In this study, we expand it to four multi-modal tracking tasks, including RGB+Depth, RGB+Thermal, RGB+Event, and RGB+Language tracking, resulting in a unified multi-modal tracking framework named SeqTrackv2. Specifically, we develop a unified interface to accommodate various auxiliary modalities. All modalities are converted into a unified sequence format and then inserted into the encoder through this interface. Furthermore, we employ a set of task-prompt tokens to specify the task. Each task-prompt token corresponds to a specific multi-modal tracking task. The task-prompt token corresponding to the current task is input into both the encoder and decoder, helping the model to process the modality associated with the current task. In this manner, SeqTrackv2 can manage diverse multi-modal tracking tasks using a unified model and parameter set, obviating the necessity to design and train specific models for each task.

Experiments demonstrate that our SeqTrack method is effective, achieving new state-of-the-art performance on several tracking benchmarks. For example, SeqTrack-B256 attains a 74.7\% AO score on GOT-10k \cite{GOT10K}, surpassing the recent OSTrack-256 tracker \cite{ostrack} by 3.7\% under aligned settings, \emph{i.e.}, using the same encoder architecture and input resolution.
Furthermore, compared to the recent state-of-the-art tracker MixFormer~\cite{mixformer}, SeqTrack-B256 operates 1.4 times faster (40 \emph{v.s.} 29 \emph{fps}) while achieving 0.7\% superior AUC score on LaSOT~\cite{LaSOT}. 
It is noteworthy that all these prior methods heavily rely on well-designed head networks and the corresponding complicated loss functions \cite{GIoU,focal_loss}. In contrast, our SeqTrack employs only a plain encoder-decoder transformer architecture with a simple cross-entropy loss.
On the multi-modal tracking tasks, SeqTrackv2 also delivers promising results. For instance, SeqTrackv2-L384 achieves a 61.0\% AUC score on the RGB+Thermal benchmark LasHeR~\cite{lasher},  74.8\% EAO score on the RGB+Depth benchmark VOT-RGBD22~\cite{vot2022}, and 63.4\% AUC score on the RGB+Event benchmark VisEvent~\cite{visevent}, outperforming ViPT~\cite{vipt} by an average of 5.1\% points. On the RGB+Language benchmark TNL2K~\cite{TNL2K}, SeqTrackv2-L384 obtains a 62.4\% AUC score, surpassing the recent JointNLT tracker~\cite{JointNLT} by 5.5\%. Notably, all these prior methods train different models for each multi-modal task, while our SeqTrackv2 employs a unified model and parameter set.

In summary, the contributions of this work are as follows:
\begin{itemize}[leftmargin=0.468cm]
\vspace{-1mm}
    \item{We propose a sequence-to-sequence learning method for visual tracking, which conceptualizes tracking as a generation task, providing a new perspective on tracking modeling.}
    \item{We propose a unified sequence-to-sequence learning method for multi-modal visual tracking. It unifies various multi-modal visual object tracking tasks with a single model and parameter set, obviating the necessity to design and train specific models for each task.}
    \item{We present a new family of sequence tracking models that strike a balance between speed and accuracy. Experiments verify the effectiveness of these new models.}
\end{itemize}

A preliminary version of this work~\cite{seqtrack} was published at the IEEE CVPR2023 conference. In this new version, we significantly extends it in the following aspects: First, we expand our previous RGB-based tracker to multi-modal tracking tasks, including RGB+Depth, RGB+Thermal, RGB+Event, and RGB+Language tracking tasks. This introduces a new sequence-to-sequence tracking method for multi-modal tracking named SeqTrackv2. Second, we propose a unified interface for various modalities and employ a set of task-prompt tokens, enabling our method to execute different multi-modal tracking tasks with a unified model and parameter set. Third, we develop a new family of multi-modal sequence tracking models that establish new state-of-the-art performance on various multi-modal tracking benchmarks. Furthermore, we conduct extensive experiments of the new SeqTrackv2 tracker and provide additional ablation studies of the previous RGB-based tracker SeqTrack.

\section{Related Work}
\textit{Visual Tracking.}
Existing tracking approaches commonly adopt a divide-and-conquer strategy, decomposing tracking into multiple subtasks. They first extract visual features from video frames using a deep neural networks~\cite{ResNet, 2017Attention, ViT}, and then design multiple task-specific head networks to predict the bounding boxes of target objects.
Based on the differences in head networks, prior trackers can be categorized into two groups: 1) trackers relying on classification and regression, and 2) trackers based on corner predictions.

Most prevalent trackers~\cite{SiameseRPN,transt,SiamFC++,ostrack,ATOM, DiMP} belong to the first category, which models tracking with a classification head for foreground-background prediction and a regression head for object scale estimation. 
For the classification head network, most trackers adopt stacked convolutional layers with various loss functions, including cross-entropy loss
\cite{SiameseRPN,transt},
focal loss~\cite{SiamFC++,ostrack}, modified $\ell_2$ loss
\cite{ATOM, DiMP}, and KL-divergence loss~\cite{PrDiMP}.
For the regression head, Siamese trackers and some transformer-based trackers adopt stacked convolutional layers with $\ell_1$ loss \cite{SiameseRPN,SiamRPNplusplus} and IoU loss\cite{transt, sbt}, while discriminative trackers\cite{ATOM, DiMP} employ the IoU-Net~\cite{IOU-Net} with MSE loss \cite{ATOM}.

STARK~\cite{Stark} and its follow-up works~\cite{CSWinTT,mixformer,simtrack} belong to the corner prediction category, which locates target objects using corner prediction heads. They employ a two-branch network to generate two probability maps for the top-left and bottom-right corners of the target object.  The final object bounding boxes are obtained by calculating the expectation of corners' probability distribution.  The loss function is a combination of $\ell_1$ and generalized IoU~\cite{GIoU} loss.

The intricate design of head networks and loss functions complicates existing tracking frameworks and exacerbates training challenges. In contrast, our method reframes tracking as a sequence generation task, eliminating the need for complicated head networks and redundant loss functions. It relies solely on a single cross-entropy loss with a plain transformer.

\textit{Multi-Modal Tracking.}
Recent advancements in single-modal visual object tracking, leveraging RGB image inputs, have yielded impressive results. Nevertheless, in complex environments or specific scenarios, relying solely on RGB imagery may not offer sufficient information to achieve accurate tracking. To address these challenges, many researchers~\cite{ca3dms, apfnet, jmmac, cmpp, JointNLT, DecoupleTNL, SNLT, protrack, vipt} have turned their attention to multi-modal tracking, which integrates additional modalities alongside RGB inputs to enhance tracking performance. 

Specifically, depth information~\cite{rgbd1k, depthtrack} provides insights into the three-dimensional position and shape of targets, thereby enhancing tracker robustness against occlusions and complex backgrounds. Thermal infrared cues~\cite{lasher, rgbt234} are crucial for tracking in low-light or adverse weather conditions, as they provide target features unaffected by environmental lighting variations. Event data~\cite{visevent, COESOT}, captured at high frame rates, offer high dynamic range and low latency perception capabilities, suitable for scenarios with rapid movements or high-speed changes. The language modality~\cite{TNL2K, TNLS}, on the other hand, can describe target attributes and behaviors through natural language, further augmenting the tracker's understanding and reasoning capabilities.

Despite the benefits, existing multi-modal tracking methods heavily rely on modality-specific design and training, resulting in distinct models and parameter sets for each multi-modal tracking task. This fragmented situation introduces complexity, requires additional computational resources, and limits the ability to leverage shared information across modalities, ultimately hindering scalability and efficiency. In contrast, our method unifies different multi-modal tracking task in a sequence model, eliminating the need for modality-specific design or training.

\textit{Sequence Learning.}
Sequence-to-sequence learning, originally proposed for natural language modeling \cite{sutskever2014sequence, cho2014learning}, has recently found applications in computer vision. 
Pix2Seq \cite{pix2seq} is a representative work that frames object detection as a token generation task conditioned on the observed pixel inputs. This sequence learning approach has also been effectively extended to other vision tasks, including instance segmentation and keypoint detection \cite{pix2seqv2}. Moreover, in cross-modality domain, sequence learning is becoming increasingly popular. For instance, text-to-image generation models such as DALL-E \cite{dalle} and vision-language models like Flamingo \cite{flamingo} all leverage sequence-to-sequence learning to unify multi-modality pretraining.

Our sequence learning framework shares a similar spirit with  Pix2Seq \cite{pix2seq}, both casting vision tasks as sequence generation problems and discretizing continuous values of bounding box coordinates into integers. However, our method differs from Pix2Seq in three fundamental ways. 
1) The construction of the sequences differs. Pix2Seq uses object corner coordinates and object categories to set up the sequence, while our method employs center point coordinates and object scales.  
2) The architectures are different. Pix2Seq adopts ResNet~\cite{ResNet} as its backbone network followed by an encoder-decoder transformer. In contrast, our method is more compact, utilizing a single encoder-decoder transformer. It employs ViT~\cite{ViT} as the encoder for feature extraction and causal transformer blocks as the decoder for sequence generation.
3) The tasks are different, Pix2Seq is designed for detection, while ours is for tracking.
Some previous tracking designs, such as online template update, can be seamlessly integrated into our method. 
4) The modalities are different, Pix2Seq is developed for RGB-based data, while ours can process auxiliary modalities including depth, thermal, event, and natural language.

\section{Seqtrack}
\label{sec:SeqTrack}
This section presents a comprehensive description of the proposed RGB-based tracking method, SeqTrack.
We start with a concise overview of our sequence-to-sequence tracking framework. Following that, we detail image and sequence representations, along with the proposed model architecture. Finally, we elucidate the training and inference pipelines, as well as the integration of tracking prior knowledge.

\begin{figure*}[htbp]
\begin{center}
\includegraphics[width=0.85\linewidth]{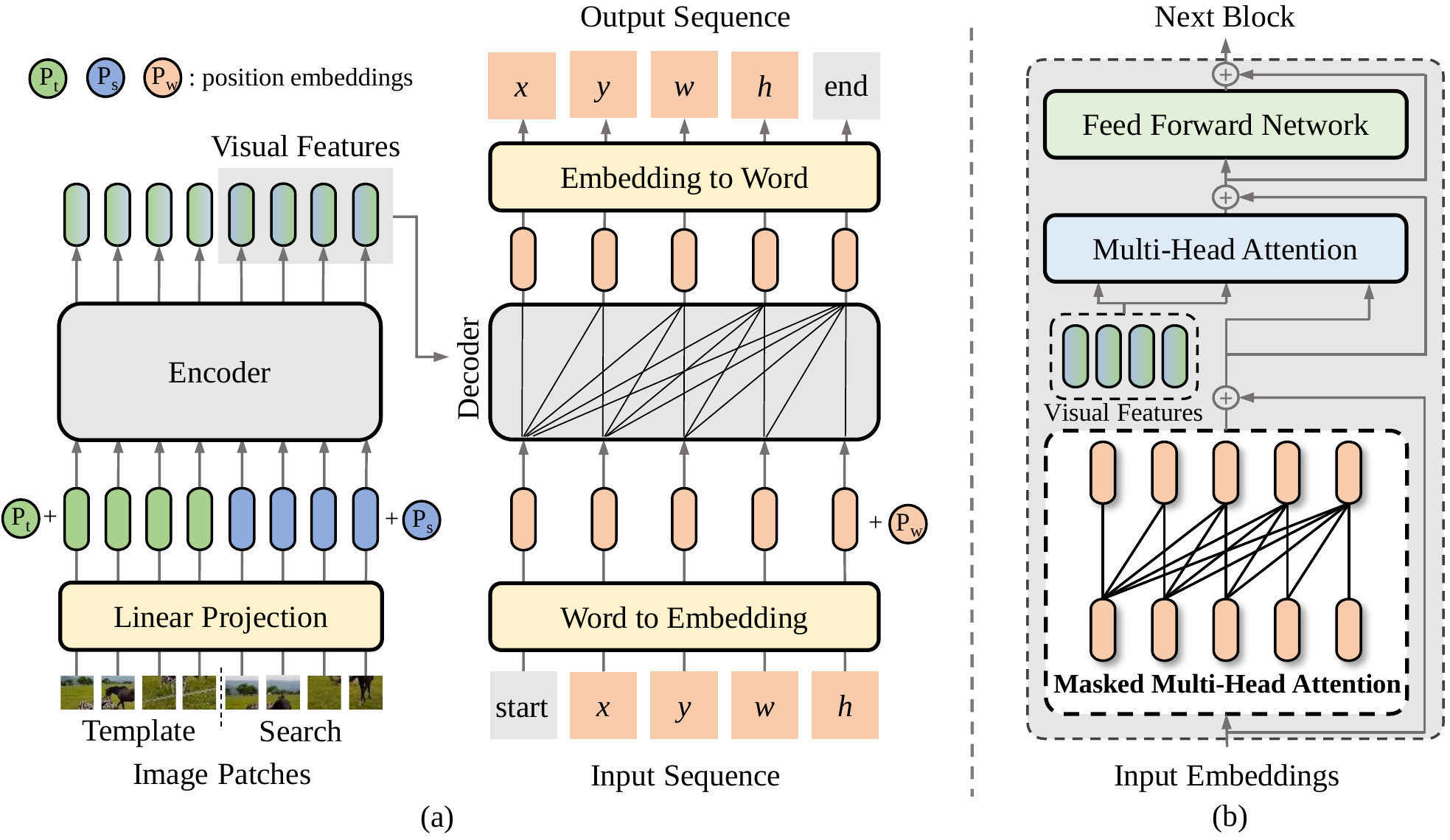}
\vspace{-3mm}
\end{center}
   \caption{(a) Architecture of the proposed SeqTrack. The key component is an encoder-decoder transformer. The encoder extracts visual features from input video frames. The causal decoder autoregressively generates the sequence of the bounding box tokens using the extracted features. (b) Detailed transformer block in the causal decoder. The input embeddings interact in a casual manner through a masked multi-head attention mechanism. The visual feature is incorporated into the decoder via a multi-head attention layer. }
\label{fig:framework}
\end{figure*}

\subsection{Overview}
\label{subsec:overview}

The overall framework of SeqTrack is depicted in Fig.~\ref{fig:framework}(a). It employs a straightforward encoder-decoder transformer architecture. The object bounding box is first converted into a sequence of discrete tokens, \emph{i.e.}, $[x,$$y,$$w,$$h]$.
The encoder extracts visual features from input video frames, while the decoder generates the sequence of bounding box tokens autoregressively using the extracted features. A causal attention mask is imposed on the self-attention modules in the decoder, restricting tokens to attend only to their preceding tokens.
In addition to the four bounding box tokens, we also employ two special tokens: \texttt{start} and \texttt{end}.
The \texttt{start} token tells the model to begin the generation, while the \texttt{end} token represents the completion of the generation.
During training, the input sequence of the decoder is $[\texttt{start},$$x,$$y,$$w,$$h]$, and the target sequence is $[x,$$y,$$w,$$h,$$\texttt{end}]$.
During inference, the decoder's input sequence initially contains a single \texttt{start} token.
At each step, a new bounding box token is generated and appended to the input sequence to produce the next one. The process continues iteratively until the four token values of the bounding box are generated, indicating the completion of the prediction.

\subsection{Image and Sequence Representation}
\label{subsec:representation}

\textit{Image Representation.}
The encoder takes as input a template image $\bm{t} \in {\mathbb{R}}^{3\times{H}\times{W}}$ and a search image $\bm{s} \in {\mathbb{R}}^{3\times{H}\times{W}}$. 
The image $\bm{t}$ represents the object of interest, while $\bm{s}$ represents the search region in subsequent video frames. In existing trackers \cite{SiameseFC, SiameseRPN, transt, Stark}, the resolution of template images is typically smaller than that of search images. 
In contrast, we maintain the same size for both the template and search images. This decision is based on the observation that including more background in the template image can enhance tracking performance.
The search and template images are partitioned into patches: $\bm{s}_p \in {\mathbb{R}}^{N \times {P^2 \times 3}}$ and $\bm{t}_p \in {\mathbb{R}}^{N \times {P^2 \times 3}}$, where $(P,$$P)$ represents the patch size, and $N$$=$$HW$$/$$P^2$ indicates the patch number. Subsequently, a linear projection is applied to map the image patches to visual embeddings. 
Learnable position embeddings \cite{2017Attention} are added to the patch embeddings to preserve positional information. These combined embeddings are then processed by the encoder.

\textit{Sequence Representation.}
We represent the target bounding box as a sequence of discrete tokens.
Specifically, a bounding box is defined by its center point $[x,$$y]$ and scale $[w,$$h]$.
There are several bounding box formats, such as $[x,$$y,$$w,$$h]$ and $[w,$$h,$$x,$$y]$.
We opt for the format $[x,$$y,$$w,$$h]$, because it aligns with human prior knowledge: first localizing object position $[x,$$y]$, and then estimating its scale $[w,$$h]$.
Each continuous coordinate is uniformly discretized into an integer within the range $[1,$ $n_{bins}]$.
We utilize a shared vocabulary $\bm{V}$ for all coordinates. Each integer between $[1,$ $n_{bins}]$ corresponds to a word in $\bm{V}$, resulting in a vocabulary size of $n_{bins}$ ($4,000$ in our experiments). The final input sequence is $[\texttt{start},$$x,$$y,$$w,$$h]$, and the target sequence is $[x,$$y,$$w,$$h,$$\texttt{end}]$.
Each word in the vocabulary $\bm{V}$ corresponds to a learnable embedding, which is optimized during training.
The special token \texttt{start} also has a corresponding learnable embedding.
The embeddings corresponding to the input words are fed into the decoder.
Since the transformer is permutation-invariant, we augment word embeddings with learnable position embeddings \cite{2017Attention}. For the final model outputs, we need to map the embeddings back to words. To accomplish this, we employ a multi-layer perceptron with a softmax function to sample words from $\bm{V}$ based on the output embeddings.

\begin{figure}[t]
\begin{center}
\includegraphics[width=1\linewidth]{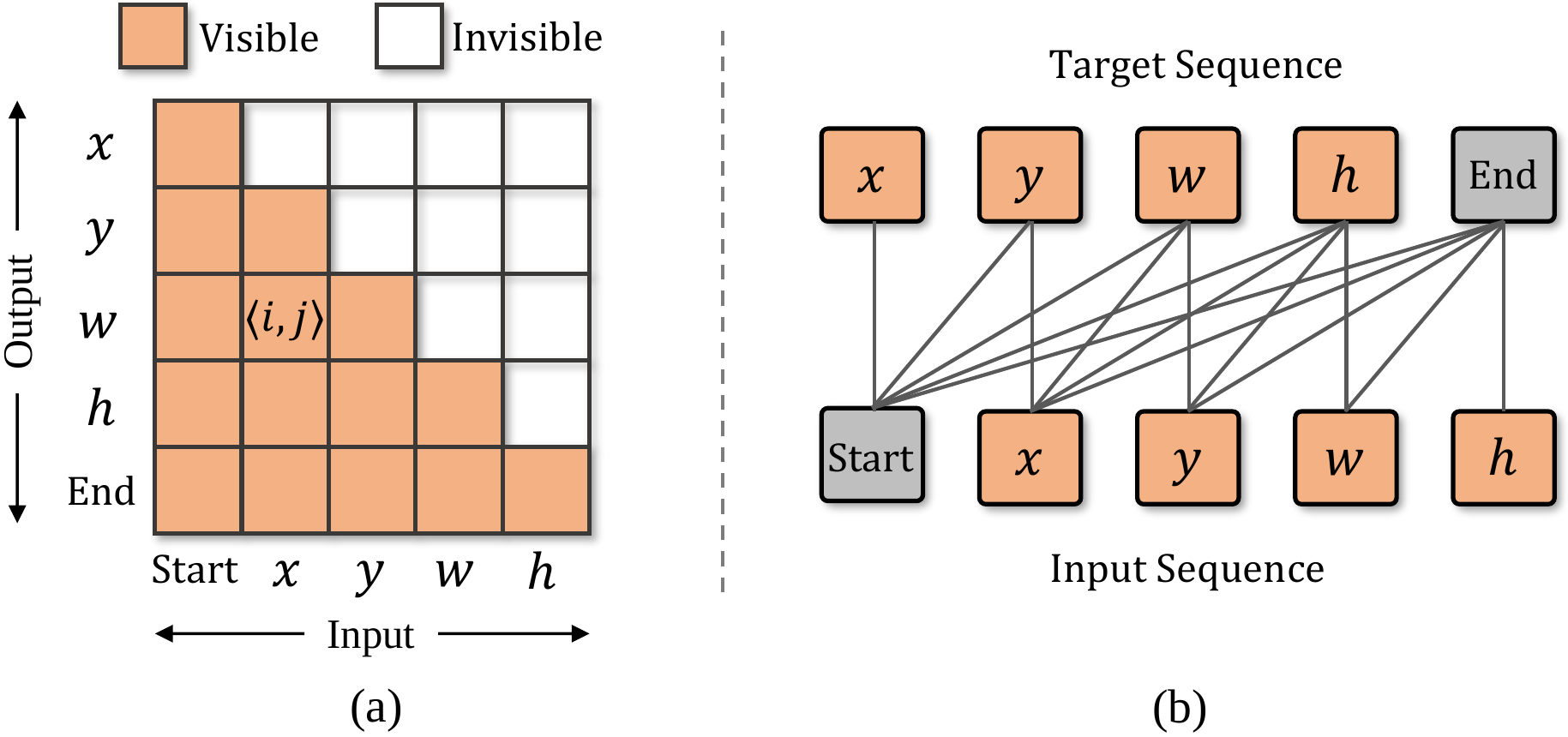}
\end{center}
\vspace{-3mm}
   \caption{(a) Illustration of the causal attention mask in the decoder using a maximum sequence length of 5 tokens. 
   An orange cell at row $i$ and column $j$ indicates that the attention mechanism is allowed to attend to the $j$th input token, when producing the $i$th output token. 
   (b) Illustration of the input and target sequences. Similar to autoregressive language modeling \cite{2017Attention}, the input sequence is the target sequence with one position offset.}
\label{fig:attn&seq}
\end{figure}

\subsection{Model Architecture}
\label{subsec:architecture}

Our model employs a simple encoder-decoder transformer architecture, as illustrated in Fig.~\ref{fig:framework}. The encoder is responsible for extracting visual features from input video frames, while the decoder predicts the bounding boxes of target objects in an autoregressive manner. 

\textit{Encoder.}
The encoder follows a standard vision transformer architecture (ViT) \cite{ViT}, with two minor modifications: i) The \texttt{CLASS} token is removed, since it is designed for image classification task.  ii) A linear projection is appended to the last layer to align the feature dimensions of the encoder and decoder.  The encoder receives patch embeddings of template and search images, producing their corresponding visual features. Only the features of the search image are fed into the decoder. The function of the encoder is to extract the visual features of search and template images in a joint way, and learn feature-level correspondence through attention layers.

\textit{Decoder.} The decoder of SeqTrack is a causal transformer \cite{2017Attention}. As illustrated in Fig.~\ref{fig:framework}(b), each transformer block consists of a masked multi-head attention, a multi-head attention, and a feed forward network (FFN). 
More concretely, the masked multi-head attention receives the word embeddings from the preceding block, and applies a causal mask to ensure that each sequence element's output depends only on its preceding elements. In other words, the attention mask restricts the output embedding at position $i$ to only attend to the input embeddings at positions before $i$, as illustrated in Fig.~\ref{fig:attn&seq}(a). After that, the multi-head attention integrates the extracted visual features into the word embeddings, which allows the word embeddings to attend to the visual features derived from the encoder. 
Finally, a feed forward network (FFN) is applied to generate embeddings for the next decoder block.

\subsection{Training and Inference}
\label{subsec:traininginference}

\textit{Training.} Similar to language modeling \cite{2017Attention}, SeqTrack is trained to maximize the log-likelihood of the target tokens conditioned on the preceding subsequence and input video frames using cross-entropy loss. The learning objective function is formulated as:
\vspace{-2mm}
\begin{equation}
\vspace{-3mm}
\begin{split}
\label{equation:objective}
{\rm{maximize}} \sum_{j=1}^{L}{\rm{log}}Q(\hat{\bm{z}}_j|\bm{s},\bm{t},\hat{\bm{z}}_{< j}), 
\end{split}
\end{equation}
where $Q(\cdot)$ represents the softmax probability, $\bm{s}$ denotes the search image, $\bm{t}$ is the template, $\hat{\bm{z}}$ is the target sequence, $j$ is the position of the token, and $L$ is the length of the target sequence.
Here, $\hat{\bm{z}}_{< j}$ represents the preceding subsequence used to predict the current token $\hat{\bm{z}}_j$.
The input sequence is the target sequence with a one-position offset (excluding \texttt{start} and \texttt{end}), as visualized in Fig.~\ref{fig:attn&seq}(b). 
Such an offset, combined with the causal masking, ensures the autoregressive property of the sequence model. 
The target sequence can be regarded as a description of the object's bounding box.
The training process aims to teach the model to ``read out" the words of the description based on the preceding words.

\textit{Inference.}
During inference, the encoder perceives the template image and the search region in successive video frames.
The initial input to the decoder is the \texttt{start} token, which signals the model to start the generation process.
Then, the model ``reads out" the target sequence $[x,$$y,$$w,$$h,$$\texttt{end}]$ token by token.
For each token, the model samples it from the vocabulary $\bm{V}$ based on maximum likelihood, \emph{i.e.}, 
$\hat{\bm{z}}_{j}$=${\argmax}_{\bm{z}_{j}}
Q(\bm{z}_{j}|\bm{s},\bm{t},\hat{\bm{z}}_{< j})$, where ${\bm{z}}_{j}$ represents the words in $\bm{V}$.
Additionally, we introduce online template update and window penalty during inference to integrate prior knowledge,  further enhancing the model's accuracy and robustness. Detailed descriptions are provided in the following subsection.

\subsection{Prior Knowledge Integration}
\label{subsec:prior}

Prior knowledge, such as window penalty \cite{SiameseFC,SiameseRPN} and online update \cite{ATOM,Stark,li2013survey}, has been widely incorporated into existing tracking models and proven effective \cite{transt,ostrack,mixformer,VITAL}. In this subsection, we discuss how to integrate such prior knowledge into the proposed sequence-to-sequence learning framework to further enhance tracking performance.

\textit{Online Update.}
Since the appearance of a target object may change dramatically during online tracking, relying solely on the initial template image may not always yield accurate tracking results.
To address this issue, we incorporate online template update \cite{Stark} into our method. More specifically, we introduce a dynamic template alongside the initial template image to capture the appearance changes of target objects. 
The dynamic template is updated on the fly. It is well recognized that poor-quality templates may lead to inferior tracking performance~\cite{mixformer}.
As a consequence, we utilize the likelihood of the generated tokens to automatically select reliable dynamic templates. 
Specifically, we compute the average softmax scores over the four generated bounding box values. If the averaged score exceeds a specific threshold $\tau$ and the update interval $T_u$ is reached, the dynamic template will be updated with the tracking result in the current frame; otherwise, it maintains the previous state. Experiments demonstrate that this simple approach can improve tracking accuracy (see the ablation in Sec.~\ref{subsec:ablation}). 
Moreover, unlike previous methods~\cite{Stark,mixformer}, our approach does not require an additional score head to determine whether to update the template, which typically necessitates a second stage of training.

\textit{Window Penalty.}
It is empirically validated that the pixel displacement between consecutive frames is relatively small \cite{SiameseFC,SiameseRPN}.
To penalize large displacements, we introduce a new window penalty strategy to our method during online inference.
Specifically, the position of the target object in the previous frame corresponds to the center point of the current search region.
The discrete coordinates of the center point in the current search region are $[\frac{n_{bins}}{2},$$\frac{n_{bins}}{2}]$.
We penalize the likelihood of integers (\emph{i.e.}, words) in the vocabulary $\bm{V}$ based on their difference to $\frac{n_{bins}}{2}$, when generating $x$ and $y$.
A large difference between an integer and $\frac{n_{bins}}{2}$ will incur a corresponding large penalty.
In implementation, the softmax scores of integers form a vector of size $n_{bins}$.
We simply multiply this vector by the Hanning window of the same size.
This effectively suppresses large displacements.
Unlike previous practice \cite{SiameseRPN,transt}, we avoid introducing of additional hyperparameters to tune the penalty magnitude.

\vspace{-1mm}

\section{Seqtrackv2}
\label{sec:seqtrackv2}

This section presents the proposed multi-modal tracking method, SeqTrackv2, in detail. First, we provide a brief overview of our  sequence-to-sequence multi-modal tracking framework. Then, we elaborate on the unified interface for auxiliary modalities and the methodology for task instruction. Finally, we describe the training and inference process.

\begin{figure*}[htbp]
\begin{center}
\includegraphics[width=0.85\linewidth]{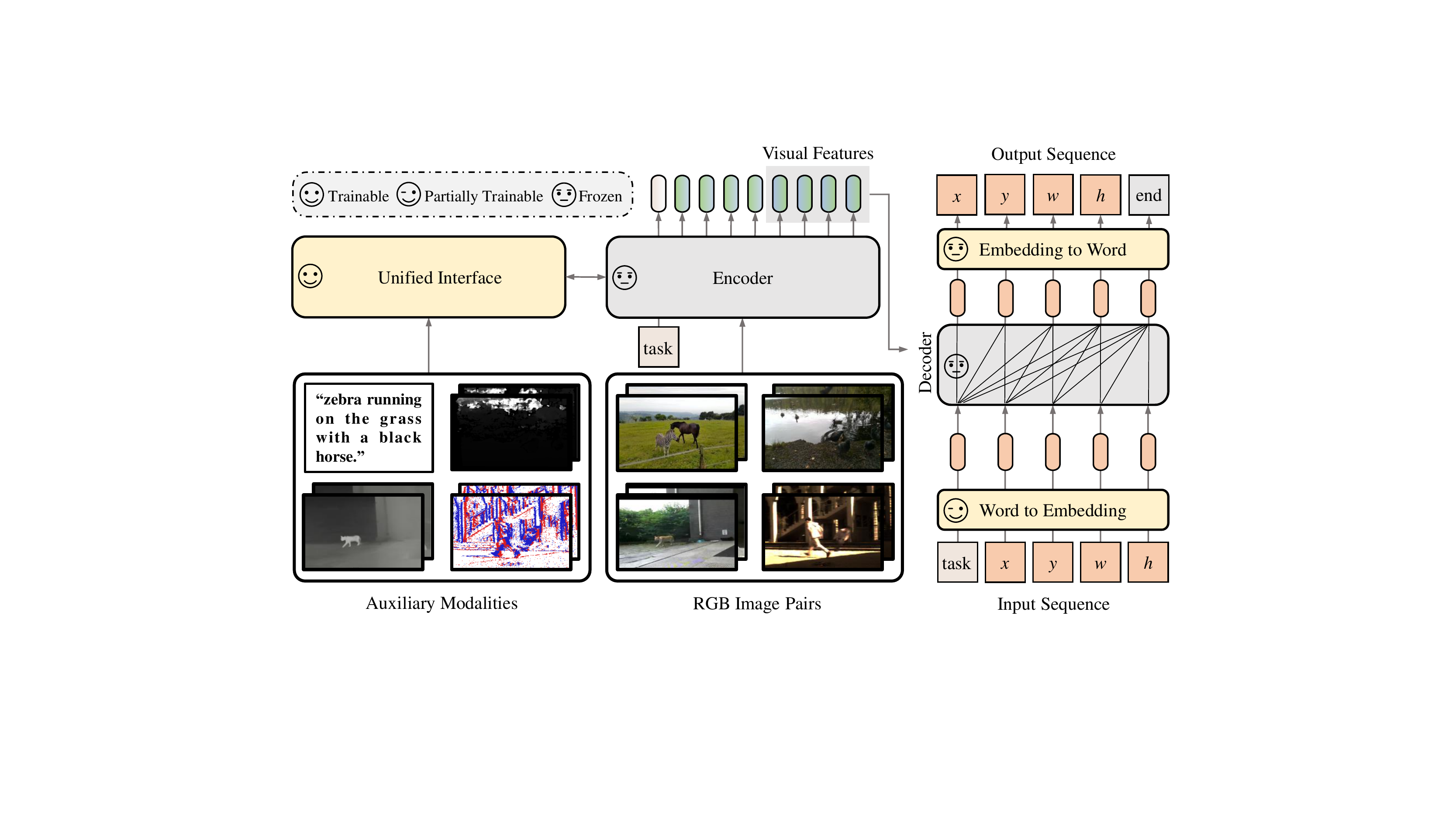}
\end{center}
\vspace{-3mm}
   \caption{Architecture of the proposed SeqTrackv2. Expanding on SeqTrack, SeqTrackv2 integrates a unified interface and task-prompt tokens. This integration consolidates diverse multi-modal tracking tasks into a unified model and parameter set. }
\vspace{-1mm}
\label{fig:frameworkv2}
\end{figure*}

\subsection{Overview}
The overall framework of the proposed SeqTrackv2 is illustrated in Fig.~\ref{fig:frameworkv2}. SeqTrackv2 extends the capabilities of its predecessor, SeqTrack, by accommodating various auxiliary modalities such as depth, thermal, event, and language inputs. Unlike traditional multi-modal tracking methods, which rely on separate models and parameter sets for each modality, SeqTrackv2 adopts a unified approach, allowing seamlessly integration of different modalities into a single model. This unification is facilitated by a unified interface and task-prompt tokens. The unified interface standardizes various modalities into a unified sequence format and feeds them into the encoder. Meanwhile, task-prompt tokens highlight the current multi-modal task, guiding the model's processing. While these enhancements augment SeqTrack's functionality, the core components and principles remain consistent across SeqTrackv2, ensuring the continuity and compatibility within our sequence tracking framework.

\subsection{Unified Interface}
\label{sec-interface}

\begin{figure}[t]
\begin{center}
\includegraphics[width=0.8\linewidth]{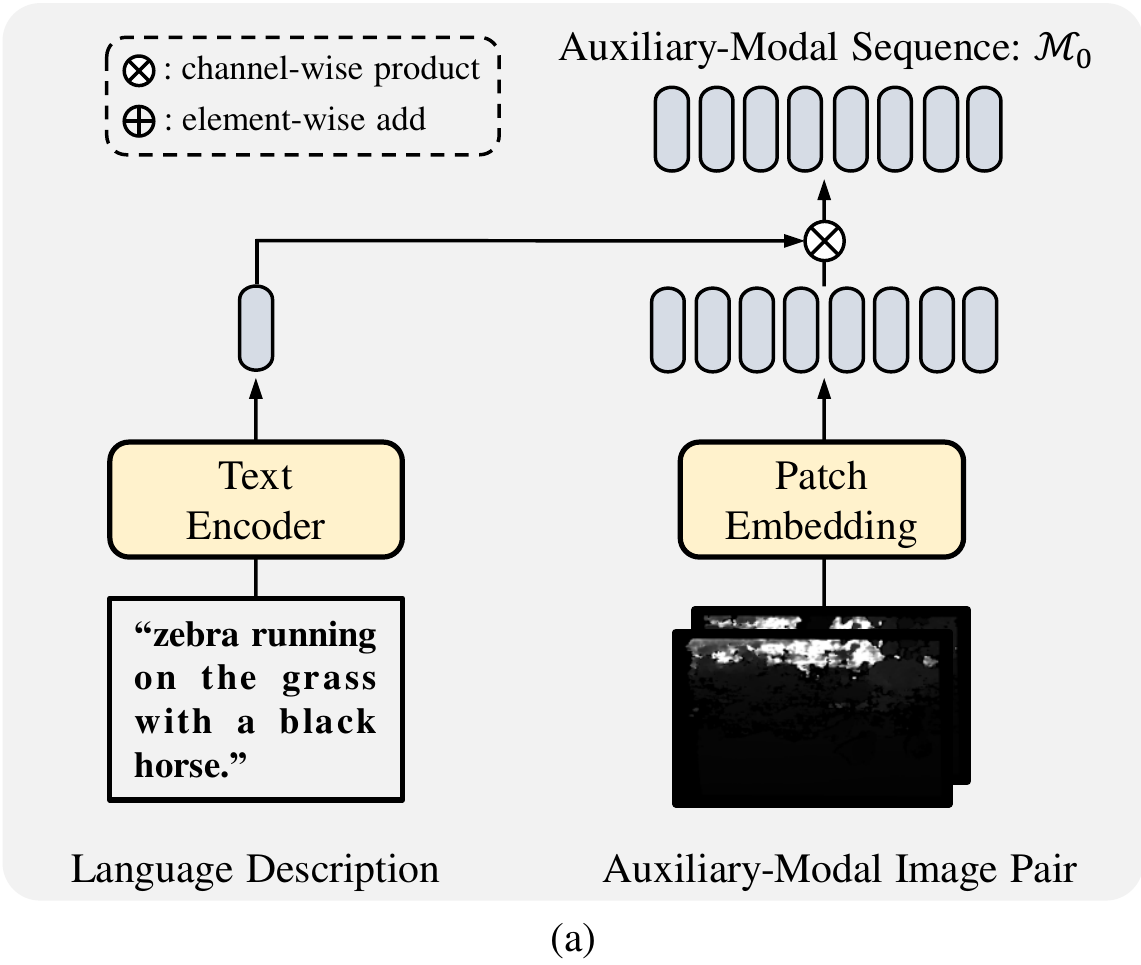}
\end{center}
\begin{center}
\includegraphics[width=0.8\linewidth]{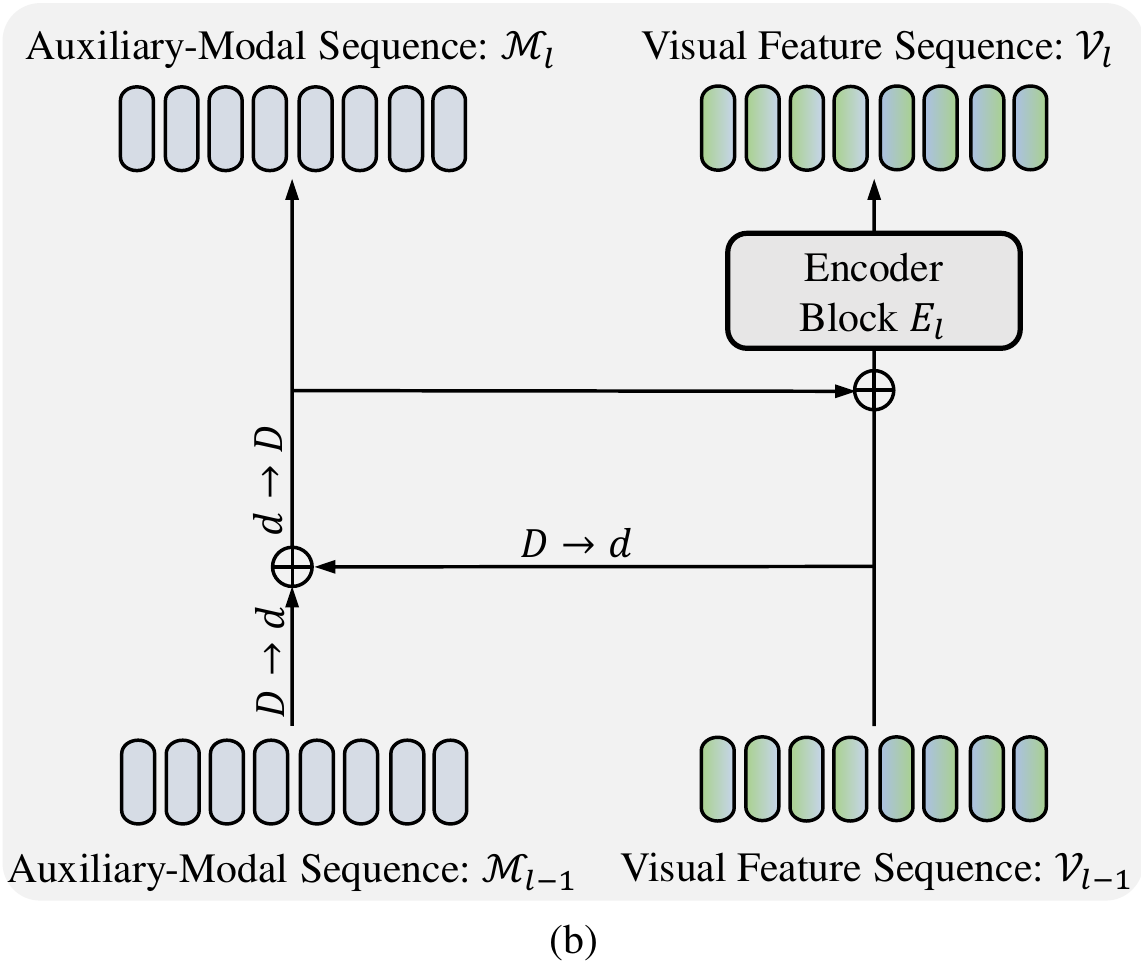}
\end{center}
    \vspace{-3mm}
   \caption{
   Details of the proposed unified interface. (a) Various auxiliary modalities are convert into a sequence format. (b) The auxiliary-modal sequence is integrated into the visual features of the encoder.} 
       \vspace{-1mm}
\label{fig:interface}
\end{figure}

We propose a unified interface for perceiving auxiliary modalities.
Initially, we standardize various modalities into a unified sequence format, as depicted in Fig.~\ref{fig:interface}(a). Specifically, for a modality with image format such as depth, thermal, and event data, we employ the standard patch embedding method~\cite{ViT} to convert it into a sequence of embeddings. Regarding the language modality, we utilize a text encoder (BERT~\cite{BERT} in our implementation) to extract the features and then apply max-pooling to obtain a single text embedding. Subsequently, we perform channel-wise product operations between the text embedding and the image-format modality's embeddings, resulting in the auxiliary-modal sequence denoted as $\mathcal{M}_0$. Through this process, various modalities are unified into a sequence format. It is noteworthy that existing multi-modal benchmarks typically provide only one kind of auxiliary modality for each data sample. Therefore, we employ the \texttt{PAD} token to fill the language sequence for samples without a language description. Similarly, for samples lacking an image-format auxiliary modality, we substitute the original RGB-based image to serve as the image-format auxiliary modality.

After obtaining the auxiliary-modal sequence, we integrate it into the visual features of the encoder. Integration is performed at every encoder block,  as illustrated in Fig.~\ref{fig:interface}(b).
Given that auxiliary modalities typically contain less information compared to RGB-based images, we integrate them in a low-rank manner for efficiency. Initially, both the auxiliary-modal sequence $\mathcal{M}_{l-1}$ and the visual feature sequence $\mathcal{V}_{l-1}$ of the encoder are projected from dimension $D$ to a lower dimension $d$ in terms of channels. Subsequently, they are element-wise added together and then projected back to dimension $D$, resulting in the new auxiliary-modal sequence $\mathcal{M}_{l}$. The new $\mathcal{M}_{l}$ is then added back to $\mathcal{V}_{l-1}$. Finally, the encoder block $\mathcal{E}_{l}$ processes this combined sequence to generate the new visual feature sequence $\mathcal{V}_{l}$. This integration effectively incorporates the information of the auxiliary modality into the visual features.

\subsection{Task Prompting}
\label{sec-taskprompt}

Although the auxiliary-modal inputs contain certain task-specific information, we have found that adding a one-token task prompt could further improve the model. Specifically, each multi-modal tracking task corresponds to a unique task-prompt token. In the encoder, we map the task-prompt token to an embedding and concatenate it with the visual embeddings before inputting it into the encoder. In the decoder, we simply replace the \texttt{START} token with the task-prompt token. This way, the task-prompt token provides information about the current task, helping the model to better adapt to the specific multi-modal tracking task at hand.

\subsection{Training and Inference}
\label{sec-train}
The training and inference  procedures of SeqTrackv2 remain essentially consistent with those of SeqTrack. Additionally, during training, SeqTrackv2 loads the parameters of SeqTrack and then freezes them, thereby capitalizing on the knowledge acquired from large-scale RGB-based data. Only the parameters of the unified interface and the embeddings of tasks-prompt tokens are trained to assist the frozen SeqTrack model in managing the multi-modal tracking tasks. Concerning multi-task training, we evenly mix the data from four multi-modal tasks  in each minibatch.

\section{Experiments}
\label{sec:Experiments}

\subsection{Implementation Details}
\label{subsec:Implementation Details}

\textit{SeqTrack Model.}
We develop four variants of SeqTrack models with different encoder architectures and input resolutions, as elaborated in Tab.~\ref{tab-model}. We adopt ViT-B~\cite{ViT} as our encoder architecture for SeqTrack-B256 and B384, and ViT-L~\cite{ViT} for SeqTrack-L256 and L384. The encoders are initialized with the MAE~\cite{mae} pre-trained parameters. The patch size is set to $16$$\times$$16$.
The decoder consists of $2$ transformer blocks, which is the same for all models. The decoder hidden size is $256$, the number of attention heads is $8$, and the hidden size of the feed forward network (FFN) is $1024$.
The number of quantization bins $n_{bins}$ and the vocabulary size are all set to $4,000$. The dimension of word embedding is $256$, which is consistent with the decoder hidden size. The embedding-to-word sampling uses a 3-layer perceptron followed by a softmax. The hidden dimension of the perceptron is $256$. The output dimension is $n_{bins}$, which is aligned with the number of words in $\bm{V}$. The word with the maximum likelihood is sampled as the output word. In addition, we present model parameters, FLOPs, and inference speed in Tab.~\ref{tab-model}. The speed is measured on Intel Core i9-9900K CPU @ 3.60GHz with 64 GB RAM and a single 2080 Ti GPU. All the models are implemented with Python 3.8 and PyTorch 1.11.0. 

\textit{SeqTrackv2 Model.}
The model details of Seqtrackv2 closely resemble those of Seqtrack, except for the addition of the new unified interface. For the unified interface, the dimension $D$ depends on the encoder's dimension, while the low-rank dimension $d$ is set to 32. The parameters of SeqTrackv2's encoder, decoder, and vocabulary are initialized with SeqTrack and subsequently frozen. In terms of the unified interface, the text encoder is initialized with BERT~\cite{BERT}. The remaining parameters are randomly initialized and made trainable.

\begin{table}[t]
\centering
\caption{Details of SeqTrack model variants.}
\label{tab-model}
\vspace{-2mm}
\setlength{\tabcolsep}{0.7mm}{
\small
\scalebox{0.95}{
\begin{tabular}{l| c c c c c}
\toprule
\multirow{2}{*}{Model}           & \multirow{2}{*}{~Encoder~} & ~Input~ & ~Params~ & ~FLOPs~ & ~Speed~ \\
~ &~ & Resolution & (M) & (G) & (\emph{fps}) \\
\midrule 
%\cmidrule(lr){1-1}\cmidrule(lr){2-6}
SeqTrack-B256~~   &   ViT-B       &   $256$$\times$$256$   &   89  &66   &  40  \\
SeqTrack-B384   &   ViT-B       &   $384$$\times$$384$   &   89  &148   &  15  \\
SeqTrack-L256   &   ViT-L      &   $256$$\times$$256$   &   309 &232    &  15  \\
SeqTrack-L384   &   ViT-L     &   $384$$\times$$384$   &   309 &524    &  5  \\
\bottomrule
\end{tabular}}
}
\vspace{-2mm}
\end{table}

\begin{table*}
  \centering
  \caption{State-of-the-art comparisons on four large-scale benchmarks. We add a symbol * over GOT-10k to indicate that the corresponding models are only trained with the GOT-10k training set. Otherwise, the models are trained with all the training data presented in Sec. \ref{subsec:Implementation Details}.
  \vspace{-2.5mm}
  The top three results are highlight with \textbf{\textcolor{cRed}{red}}, \textcolor{blue}{blue} and \textcolor{cGreen}{green} fonts, respectively.}
  \label{tab-sota}
\resizebox{1\linewidth}{!}{
  \setlength{\tabcolsep}{2mm}{  
  \small
  \begin{tabular}{l|l| ccc c ccc c ccc c ccc}
    \toprule
    &\multirow{2}*{Method} & \multicolumn{3}{c}{LaSOT~\cite{LaSOT}}&& \multicolumn{3}{c}{LaSOT$_{ext}$~\cite{lasot_journal}} && \multicolumn{3}{c}{TrackingNet~\cite{trackingnet}} && \multicolumn{3}{c}{GOT-10k*~\cite{GOT10K}}\\
    \cline{3-5}
    \cline{7-9}
    \cline{11-13}
    \cline{15-17}
    && AUC&P$_{Norm}$&P && AUC&P$_{Norm}$&P && AUC&P$_{Norm}$&P && AO&SR$_{0.5}$&SR$_{0.75}$\\
    \midrule[0.5pt]
    \multirow{4}*{\rotatebox{90}{Seq2Seq}} &SeqTrack-L384	&\textbf{\textcolor{cRed}{72.5}}	&\textcolor{blue}{81.5}	&\textbf{\textcolor{cRed}{79.3}} & &\textbf{\textcolor{cRed}{50.7}} &\textbf{\textcolor{cRed}{61.6}} &\textcolor{blue}{57.5} & &\textbf{\textcolor{cRed}{85.5}}	&\textbf{\textcolor{cRed}{89.8}}	&\textbf{\textcolor{cRed}{85.8}} & &\textbf{\textcolor{cRed}{74.8}} &81.9	&\textbf{\textcolor{cRed}{72.2}} \\
    &SeqTrack-L256	&\textcolor{blue}{72.1}	&\textbf{\textcolor{cRed}{81.7}}	&\textcolor{blue}{79.0} & &\textcolor{blue}{50.5} &\textcolor{blue}{61.5} &\textcolor{cGreen}{57.2} & &\textcolor{blue}{85.0} &\textcolor{blue}{89.5}	&\textcolor{blue}{84.9}  & &\textcolor{cGreen}{74.5} &\textcolor{cGreen}{83.2} &\textcolor{blue}{72.0} \\
    &SeqTrack-B384	&\textcolor{cGreen}{71.5}	&\textcolor{cGreen}{81.1}	&\textcolor{cGreen}{77.8} & &\textcolor{blue}{50.5} &\textbf{\textcolor{cRed}{61.6}} &\textcolor{blue}{57.5} & &\textcolor{cGreen}{83.9} &88.8	&83.6	 & &\textcolor{cGreen}{74.5} &\textcolor{blue}{84.3} &71.4\\
    &SeqTrack-B256	&69.9	&79.7	&76.3 & &\textcolor{cGreen}{49.5} &60.8 &56.3 & &83.3	&88.3	&82.2  & &\textcolor{blue}{74.7} &\textbf{\textcolor{cRed}{84.7}} &\textcolor{cGreen}{71.8}\\
    \midrule[0.1pt]
    \multirow{7}*{\rotatebox{90}{Corner Prediction}} &SimTrack~\cite{simtrack}	&70.5	&79.7	&-  &  &- &- &-  &  &83.4 &87.4	&-  &  &69.8	&78.8	&66.0\\
    &Mixformer-L~\cite{mixformer}	&70.1	&79.9	&76.3 & &- &- &- & &\textcolor{cGreen}{83.9} &\textcolor{cGreen}{88.9}	&83.1 & &-	&-	&-\\
    &Mixformer-22k~\cite{mixformer}	&69.2	&78.7	&74.7 & &- &- &- & &83.1	&88.1	&81.6 & &70.7	&80.0	&67.8\\
    &AiATrack~\cite{AiATrack}	&69.0	&79.4	&73.8  &  &47.7 &55.6 &55.4  & &82.7 &87.8  &80.4  &  &69.6	&63.2	&80.0\\
    &UTT~\cite{UTT}	&64.6	&-	&67.2 & &- &- &- & &79.7 &- &77.0 & &67.2	&76.3	&60.5\\
    &CSWinTT~\cite{CSWinTT}	&66.2	&75.2	&70.9 & &- &- &- & &81.9	&86.7	&79.5 &	&69.4	&78.9	&65.4\\
    &STARK~\cite{Stark}	&67.1	&77.0	&- & &- &- &- & &82.0	&86.9	&- &	&68.8	&78.1	&64.1\\
    \midrule[0.1pt]
    \multirow{22}*{\rotatebox{90}{Classification + Regression}} 
    &OSTrack-384~\cite{ostrack}	&71.1	&\textcolor{cGreen}{81.1}	&77.6 & &\textcolor{blue}{50.5} &\textcolor{cGreen}{61.3} &\textbf{\textcolor{cRed}{57.6}} &  &\textcolor{cGreen}{83.9} &88.5	&\textcolor{cGreen}{83.2}  &  &73.7	&\textcolor{cGreen}{83.2}	&70.8\\
    &OSTrack-256~\cite{ostrack}	&69.1	&78.7	&75.2  &  &47.4 &57.3 &53.3  &  &83.1 &87.8	&82.0  &  &71.0	&80.4	&68.2\\
    &SwinTrack~\cite{swintrack}	&71.3	&-	&76.5 & &49.1 &- &55.6 & &84.0	&-	&82.8 &	&72.4	&-	&67.8\\
    &RTS~\cite{RTS}	&69.7	&76.2	&73.7  &  &- &- &-  &  &81.6 &86.0  &79.4  &  &-	&-	&-\\
    &Unicorn~\cite{unicorn}	&68.5	&-	&-  &  &- &- &-  &  &83.0 &86.4	&82.2  &  &-	&-	&-\\
    &SLT~\cite{SLT}	&66.8	&75.5	&- & &- &- &- & &82.8 &87.5	&81.4 & &67.5	&76.5	&60.3\\
    &SBT~\cite{sbt}	&66.7	&-	&71.1 & &- &- &- & &-	&-	&-	& &70.4	&80.8	&64.7\\
    &ToMP~\cite{ToMP}	&68.5	&79.2	&73.5 & &45.9 &- &- & &81.5	&86.4	&78.9 &	&-	&-	&-\\
    &KeepTrack~\cite{keeptrack}	&67.1	&77.2	&70.2 & &48.2 &- &- & &-	&-	&-	& &-	&-	&-\\
    &AutoMatch~\cite{automatch} 	&58.3	&-	&59.9 & &- &- &- & &76.0	&-	&72.6	& &65.2	&76.6	&54.3\\
    &TransT~\cite{transt}	&64.9	&73.8	&69.0 & &- &- &- & &81.4	&86.7	&80.3 &	&67.1	&76.8	&60.9\\
    &TrDiMP~\cite{TMT}  	&63.9	&-	&61.4 & &- &- &- &	&78.4	&83.3 	&73.1	&	&68.8	&80.5	&59.7\\
    &SiamAttn~\cite{SiamAtt}  	&56.0	&64.8	&- & &- &- &- &	&75.2	&81.7	&-	&	&-	&-	&-\\
    &SiamBAN~\cite{SiamBAN}  	&51.4	&59.8	&- & &- &- &- &	&-	&- 	&-	&	&-	&-	&-\\
    &DSTrpn~\cite{DSTrpn}	&43.4	&54.4	&- & &- &- &-	& &64.9	&-	&58.9 &	&-	&-	&-\\
    &Ocean~\cite{Ocean}	&56.0	&65.1	&56.6 & &- &- &-	& &-	&-	&- &	&61.1	&72.1	&47.3\\
    &SiamR-CNN~\cite{SiamRCNN}  	&64.8	&72.2	&- &	&- &- &- & &81.2	&85.4	&80.0	&	&64.9	&72.8	&59.7\\
    &DiMP~\cite{DiMP}	   	&56.9	&65.0	&56.7 &	&39.2 &47.6 &45.1 & &74.0	&80.1	&68.7 &	&61.1	&71.7	&49.2\\
    &SiamPRN++~\cite{SiamRPNplusplus}	&49.6	&56.9	&49.1 & &34.0 &41.6 &39.6 &	&73.3	&80.0	&69.4 &	&51.7	&61.6	&32.5\\
    &ATOM~\cite{ATOM}	   	&51.5	&57.6	&50.5 &	&37.6 &45.9 &43.0 & &70.3	&77.1	&64.8	& &55.6	&63.4	&40.2\\
    %&MDNet~\cite{MDNet}	   	 &39.7	&46.0	&37.3	& &27.9 &34.9 &31.8 & &60.6	&70.5	&56.5	& &29.9	&30.3	&9.9\\
  \bottomrule
\end{tabular}
}}
  \vspace{-4mm}
\end{table*}

\textit{SeqTrack Training.} 
Our training data includes the training splits of COCO~\cite{COCO}, LaSOT~\cite{LaSOT}, GOT-10k~\cite{GOT10K}, and TrackingNet~\cite{trackingnet}. Aligned with VOT2020 evaluation protocol \cite{vot2020}, we remove the $1k$ forbidden videos in GOT-10k during training. For the evaluation on GOT-10k test set, we follow the official requirements \cite{GOT10K} and only use the training split of GOT-10k. 
The template and search images are obtained by expanding the target bounding boxes by a factor of $4$. Horizontal flip and brightness jittering are used for data augmentation. We train the model with AdamW~\cite{AdamW} optimizer and set the learning rate of the encoder to $1$$e$$-5$, the decoder and remaining modules to $1$$e-$$4$, and the weight decay to $1$$e$$-4$. The training of SeqTrack are conducted on Intel Xeon CPU E5-2690 v4 @ 2.60GHz with 512 GB RAM and 8 Tesla A100 GPUs with 80GB memory. Each GPU holds $8$ image pairs, resulting in a total batch size of $64$. The model is trained for a total of $500$ epochs with $60k$ image pairs per epoch. The learning rate decreases by a factor of $10$ after $400$ epochs. 

\textit{SeqTrackv2 Training.} 
The training data of SeqTrackv2 encompasses the training sets of DepthTrack~\cite{depthtrack}, VisEvent~\cite{visevent}, LasHeR~\cite{lasher}, TNL2K~\cite{TNL2K}, RefCOCOg~\cite{refcocog}, OTB99~\cite{TNLS}, and LaSOT~\cite{lasot_journal}. The training is conducted on Intel Xeon CPU Gold 6330 @ 2.00GHz with 512 GB RAM and 4 Tesla A40 GPUs with 48GB memory. Batch sizes are configured as follows: 128, 48, 64, and 16 for SeqTrackv2-B256, -B384, -L256, and -L384, respectively. The model undergoes training for a total of 240 epochs, with 60,000 samples per epoch. The learning rate decreases by a factor of 10 after 192 epochs. All other settings remain consistent with those of SeqTrack.

\textit{Inference.}
The online template update interval $T_u$ is set to $25$ by default, while the update threshold $\tau$ is set to $0.015$. For window penalty, the softmax likelihood of the $4,000$ words in the vocabulary $\bm{V}$ are directly multiplied by a 1D Hanning window of size $4,000$.

\subsection{State-of-the-Art Comparisons of SeqTrack}
\label{subsec:sotav1}

We compare our SeqTrack with state-of-the-art trackers on eight tracking benchmarks.

\begin{figure}[t!]
\centering
    \includegraphics[width=1\linewidth]{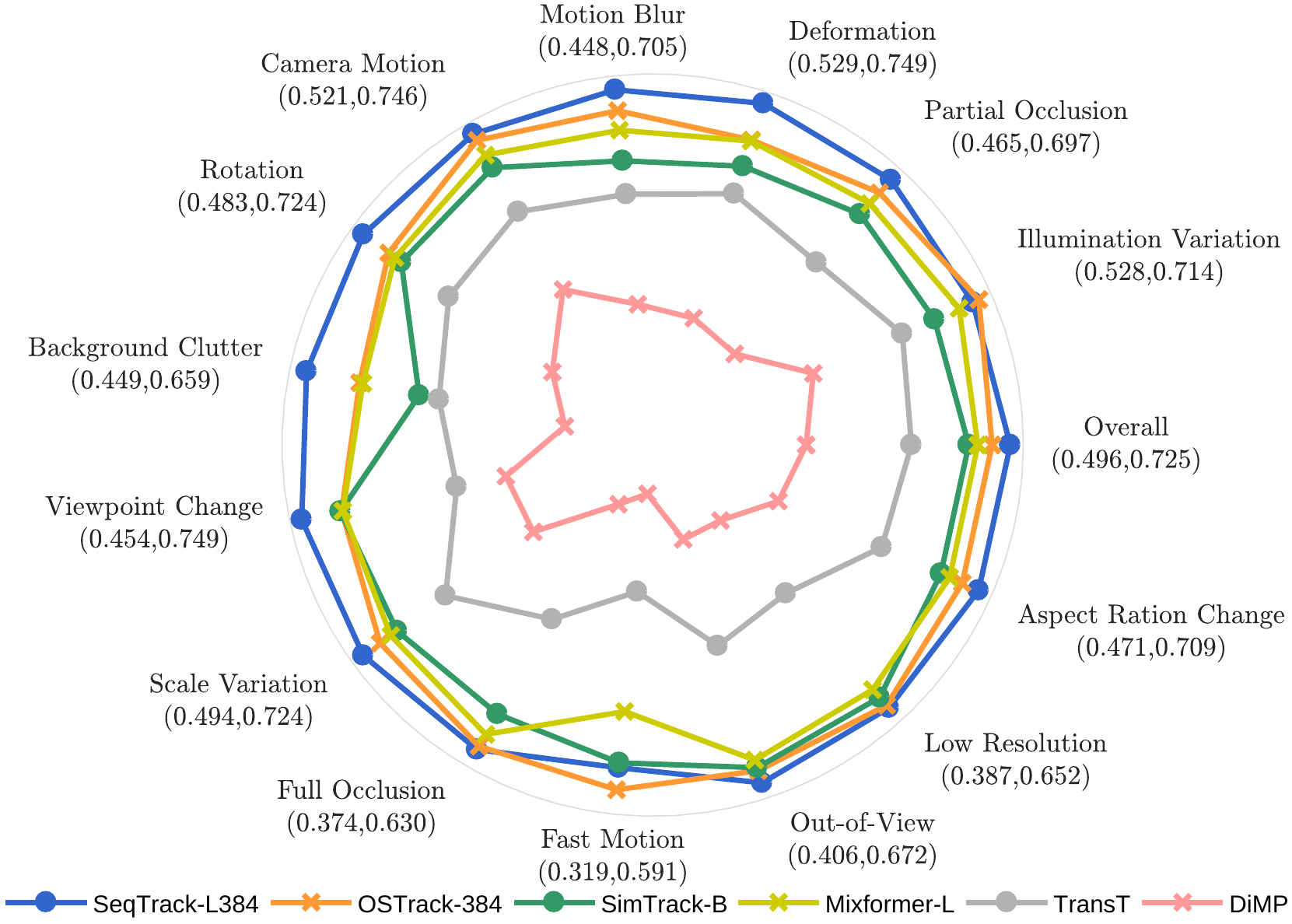}
    \vspace{-7mm}
    \caption{AUC scores of different attributes on LaSOT \cite{LaSOT}}
    \label{fig:lasotattr}
      \vspace{-1mm}
\end{figure}

\textit{LaSOT.}
LaSOT~\cite{LaSOT} is a large-scale long-term tracking benchmark. The test set comprises 280 videos, with an average length of 2448 frames. As reported in Tab.~\ref{tab-sota}, SeqTrack-B256 performs slightly better than the recent state-of-the-art method OSTrack-256 \cite{ostrack}, achieving a 0.8\% improvement in AUC score, using the same ViT-B encoder architecture and input resolution.  SeqTrack-B384 surpasses all previous trackers with an AUC score of 71.5\%.  Furthermore, SeqTrack-L384 and L256 obtain new state-of-the-art AUC scores of 72.5\% and 72.1\%, respectively. SeqTrack-L384 surpasses the previous best tracker SwinTrack \cite{swintrack} by 1.2\%.
Fig.~\ref{fig:lasotattr} illustrates the results of attribute-based evaluation, demonstrating that our SeqTrack-L384 performs better than other competing trackers on almost all attributes. In particular, our method exhibits significant advantages in the attributes of deformation and background clutter, showcasing a superior discriminative ability of the model. 

\textit{LaSOT$_{ext}$.}
LaSOT$_{ext}$~\cite{lasot_journal} is a recently released extension to the LaSOT dataset, comprising of 150 video sequences spanning 15 object classes. The results on LaSOT$_{ext}$ are aso presented in Tab.~\ref{tab-sota}.
Under aligned settings, SeqTrack-B256 achieves 2.1\% higher AUC score compared to OSTrack-256. With a more powerful ViT-L encoder, SeqTrack-L384 achieves the highest AUC score of 50.7\%.

\textit{TrackingNet.}
TrackingNet~\cite{trackingnet} is a large-scale tracking dataset. Its test set contains 511 videos covering diverse object categories and scenes. As reported in Tab.~\ref{tab-sota}, SeqTrack-B384 and SeqTrack-B256 achieve competitive results compared with previous state-of-the-art trackers. SeqTrack-L384 obtains the best AUC of 85.5\%, surpassing the previous best tracker SwinTrack by 1.5\%.

\textit{GOT-10k.}
GOT-10k~\cite{GOT10K} test set contains 180 videos covering a wide range of common challenges in tracking. Following the official requirements, we solely use the GOT-10k training set to train our models. 
As reported in Tab.~\ref{tab-sota}, SeqTrack-B256 obtains 3.7\% improvement over OSTrack-256 under aligned settings. SeqTrack-L384 achieves the best AO score of 74.8\%, surpassing the previous state-of-the-art method by 1.1\%.

\textit{TNL2K, NFS and UAV123.}
We assess our trackers on three additional benchmarks: TNL2K~\cite{TNL2K}, NFS~\cite{NFS}, and UAV123~\cite{UAV}. TNL2K is a recently introduced large-scale dataset comprising 700 challenging video sequences. NFS and UAV123 are two small-scale benchmarks including 100 and 123 videos, respectively. On the large-scale TNL2K benchmark, our SeqTrack-L384 achieves a new state-of-the-art performance with 57.8\% AUC score, as reported in Tab.~\ref{tab-sota-small}. On the small-scale benchmarks NFS and UAV123, Tab.~\ref{tab-sota-small} shows our SeqTrack models also achieve competitive results, being comparable or slightly better than the most recent trackers OSTrack and SimTrack.

\begin{figure}[t!]
\centering
\begin{center}
    \includegraphics[width=1\linewidth]{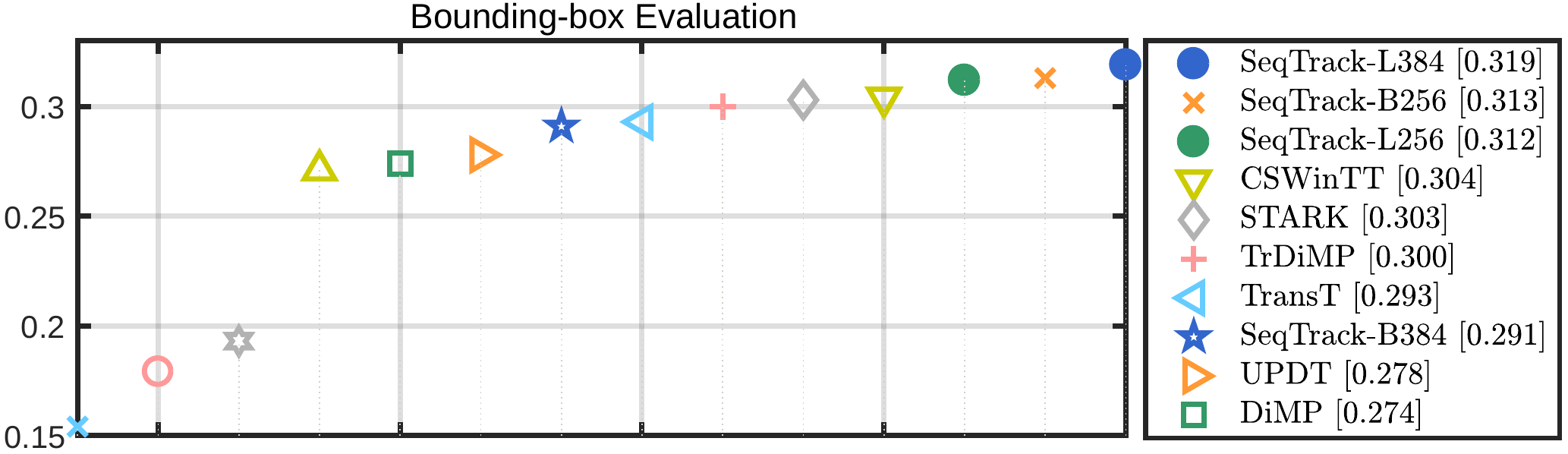}
    \vspace{-2mm}
\end{center}
\begin{center}
\vspace{-2mm}
    \hspace{-0.8mm}\includegraphics[width=1\linewidth]{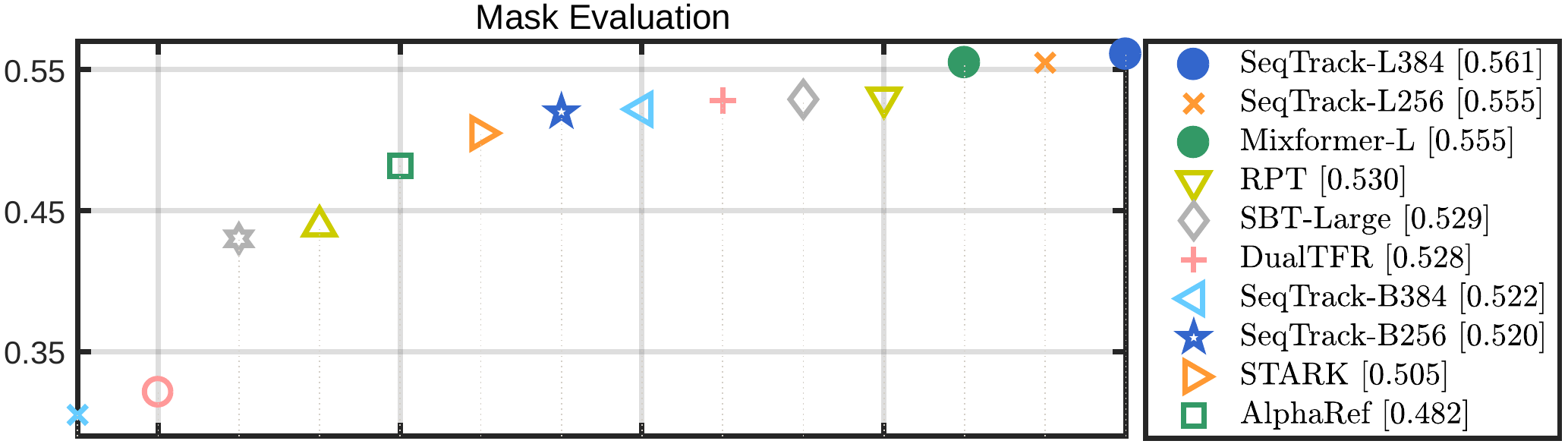}
\end{center}
\vspace{-5mm}
\caption{EAO rank plots on VOT2020. For the mask evaluation, we use Alpha-Refine~\cite{Alpha-Refine} to predict masks.}
\vspace{-2mm}
\label{fig:votrank}
\end{figure}

\textit{VOT2020.}
The VOT2020~\cite{vot2020} benchmark contains 60 challenging videos. VOT employs binary segmentation masks as ground-truth annotations. To generate segmentation masks, we equip SeqTrack with Alpha-Refine~\cite{Alpha-Refine}.
We evaluate our models by submitting both the bounding boxes and the segmentation masks.
As depicted in Fig.~\ref{fig:votrank}, our SeqTrack-L384 achieves the best results with EAO scores of 31.9\% and 56.1\% on bounding box and mask evaluations, respectively.

\begin{table}[t]\normalsize
    \vspace{-1mm}
    \caption{Comparison with state-of-the-art methods on additional benchmarks in AUC score.}
\label{tab-sota-small}
\vspace{-2mm}
  \centering
\resizebox{1\linewidth}{!}{
  \setlength{\tabcolsep}{3mm}{
    \small
    \begin{tabular}{l|ccc}
    \toprule
    Method &TNL2K~\cite{TNL2K}&NFS~\cite{NFS}&UAV123~\cite{UAV}\\
    \midrule[0.5pt]
    SeqTrack-L384 &\textbf{\textcolor{cRed}{57.8}}&66.2&68.5 \\
    SeqTrack-L256 &\textcolor{blue}{56.9}&\textcolor{blue}{66.9}&\textcolor{cGreen}{69.7} \\
    SeqTrack-B384 &\textcolor{cGreen}{56.4}&\textcolor{cGreen}{66.7}&68.6 \\
    SeqTrack-B256 &54.9&\textbf{\textcolor{cRed}{67.6}}&69.2 \\
    \midrule[0.1pt]
    OSTrack~\cite{ostrack}&55.9&66.5&\textcolor{blue}{70.7} \\
    SimTrack~\cite{simtrack}&55.6&-&\textbf{\textcolor{cRed}{71.2}} \\
    STARK~\cite{Stark}&-&66.2&68.2 \\
    TransT~\cite{transt}&50.7&65.7&69.1 \\
    TrDiMP~\cite{TMT}&-&66.5&67.5 \\
    DiMP~\cite{DiMP}&44.7&61.8&64.3 \\
    Ocean~\cite{Ocean}&38.4&49.4&57.4 \\
    ATOM~\cite{ATOM}&40.1&58.3&63.2 \\
    ECO~\cite{ECO}&32.6&52.2&53.5 \\
    RT-MDNet~\cite{RTMDNet}&-&43.3&52.8 \\
    SiamFC\cite{SiameseFC}&29.5&37.7&46.8 \\
    \bottomrule
    \end{tabular}
    }
  }
  \vspace{-2mm}
\end{table}

\subsection{State-of-the-Art Comparisons of SeqTrackv2}

We compare our SeqTrackv2 with state-of-the-art trackers on seven benchmarks across four multi-modal tracking tasks.

\textit{LasHeR.} 
LasHeR~\cite{lasher} is a comprehensive benchmark with high diversity for RGB+Thermal tracking. We evaluate our method on its 245 testing video sequences, and report the obtained AUC and Precision (P) scores in Tab.~\ref{tab-sota-rgbt}. Our four SeqTrackv2 models all surpass previous RGB+Thermal tracking methods, establishing new state-of-the-art performance. SeqTrackv2-L384 obtains the highest AUC score of 61.0 \%, outperforming the prior best trackers ViPT~\cite{vipt} by a large margin of 8.5 \%.

\textit{RGBT234.}
RGBT234~\cite{rgbt234} is a substantial RGB+Thermal tracking benchmark with 234 videos comprising visible and thermal infrared pairs. Its distribution differs from our training set, thereby enabling a more comprehensive validation of method versatility. As reported in Tab.~\ref{tab-sota-rgbt}, all four SeqTrackv2 models obtains superior performance than previous RGB+Thermal trackers. SeqTrackv2-L256 achieves the best AUC score of 68.5\%, outperforming the recent state-of-the-art tracker Un-Track~\cite{untrack} by 6.0\%. The superior performance on RGBT234 demonstrate the good versatility of our method.

\textit{VOT-RGBD22.}
VOT-RGBD2022~\cite{vot2022} is a contemporary benchmark comprising 127 RGB+Depth sequences. The evaluation protocol utilized is anchor-based~\cite{vot2020}, necessitating trackers to undergo multiple initialization starts from diverse points. The expected average overlap (EAO) serves as the main performance metric. As reported in Tab.~\ref{tab-sota-rgbd}, our  SeqTrackv2 models surpass all previous RGB+Depth tracking methods. SeqTrackv2-B384 obtains the best EAO score of 75.5\%.

\textit{DepthTrack.}
DepthTrack~\cite{depthtrack} is a comprehensive long-term RGB+Depth tracking benchmark, comprising 150 training and 50 testing videos featuring 15 per-frame attributes. The primary measure is F-score, typically used for long-term tracking. Although our method is not equipped with any re-detection module for long-term tracking, Tab.~\ref{tab-sota-rgbd} shows our SeqTrackv2-B256 still achieves a new state-of-the-art performance of 63.2\% F-score.

\textit{VisEvent.}
VisEvent~\cite{visevent} is a largest RGB+Event benchmark collected from the real world. The test set of VisEvent comprises 320 videos. As shown in Tab.~\ref{tab-sota-rgbe}, our smallest model SeqTrackv2-B256 has surpassed all previous RGB+Event tracking methods with an AUC score of 61.2\%. With the model scaling-up, our SeqTrackv2-L384 model establishes the best AUC score of 63.4\%, outperforming the previous best tracker ViPT~\cite{vipt} by 4.2\%.

\begin{table}[t]\normalsize
    \vspace{-1mm}
    \caption{State-of-the-art comparisons on RGB+Therm tracking}
\label{tab-sota-rgbt}
\vspace{-2mm}
  \centering
\resizebox{1\linewidth}{!}{
  \setlength{\tabcolsep}{4mm}{
    \small
    \begin{tabular}{l|ccccc}
    \toprule
    \multirow{2}*{Method} & \multicolumn{2}{c}{LasHeR~\cite{lasher}} & & \multicolumn{2}{c}{RGBT234~\cite{rgbt234}} \\
        \cline{2-3} \cline{5-6}
 & AUC & P & &MSR &MPR \\
    \midrule[0.5pt]
    SeqTrackv2-L384 &\textbf{\textcolor{cRed}{61.0}}&\textbf{\textcolor{cRed}{76.7}} & &\textcolor{blue}{68.0}&\textcolor{blue}{91.3}\\
    SeqTrackv2-L256 &\textcolor{blue}{58.8}&\textcolor{blue}{74.1} & &\textbf{\textcolor{cRed}{68.5}}&\textbf{\textcolor{cRed}{92.3}}\\
    SeqTrackv2-B384  &\textcolor{cGreen}{56.2}&\textcolor{cGreen}{71.5} & &\textcolor{cGreen}{66.3}&\textcolor{cGreen}{90.0}\\
    SeqTrackv2-B256 &55.8&70.4 & &64.7&88.0 \\
    \midrule[0.1pt]
    Un-Track~\cite{untrack} &-&- & &62.5&84.2\\
    ViPT~\cite{vipt} &52.5&65.1 & &61.7&83.5\\
    ProTrack~\cite{protrack} &42.0&53.8 & &59.9&79.5\\
    OSTrack~\cite{ostrack} &41.2&51.5 & &54.9&72.9\\
    %TransT~\cite{transt} &39.4&52.4 &&\\
    APFNet~\cite{apfnet} &36.2&50.0 & &57.9&82.7\\
    CMPP~\cite{cmpp} &-&- & &57.5&82.3\\
    JMMAC~\cite{jmmac} &-&- & &57.3&79.0\\
    %STARK~\cite{Stark} &36.1&44.9 &&\\
    mfDiMP~\cite{mfdimp} &34.3&44.7 & &42.8&64.6\\
    DAPNet~\cite{dapnet} &31.4&43.1 & &-&-\\
    CAT~\cite{cat} &31.4&45.0 & &56.1&80.4\\
    HMFT~\cite{vtuav} &31.3&43.6 & &-&-\\
    MaCNet~\cite{macnet} &-&- & &55.4&79.0\\
    FANet~\cite{fanet} &30.9&44.1 & &55.3&78.7\\
    DAFNet~\cite{dafnet} &-&- & &54.4&79.6 \\
    SGT~\cite{sgt} &25.1&36.5 & &47.2&72.0 \\
    %SGT++~\cite{} &25.1&36.5 & &-&-\\
    \bottomrule
    \end{tabular}
    }
  }
\end{table}

\begin{table}[t]\normalsize
    \caption{State-of-the-art comparisons on RGB+Depth tracking}
\label{tab-sota-rgbd}
\vspace{-2mm}
  \centering
\resizebox{1\linewidth}{!}{
  \setlength{\tabcolsep}{2mm}{
    \small
    \begin{tabular}{l|ccc c ccc}
    \toprule
    \multirow{2}*{Method} & \multicolumn{3}{c}{VOT-RGBD22~\cite{vot2022}} & & \multicolumn{3}{c}{DepthTrack~\cite{depthtrack}} \\
        \cline{2-4} \cline{6-8}
 & EAO & Acc. & Rob. & &F-score &Re &Pr \\
    \midrule[0.5pt]
    SeqTrackv2-L384 &\textcolor{cGreen}{74.8}&\textbf{\textcolor{cRed}{82.6}}&\textcolor{blue}{91.0} & &\textcolor{cGreen}{62.3}&\textcolor{cGreen}{62.6}&\textcolor{blue}{62.5}\\
    SeqTrackv2-L256 &\textcolor{blue}{74.9}&81.3&\textbf{\textcolor{cRed}{91.8}} & &\textcolor{blue}{62.8}&\textcolor{blue}{63.0}&\textcolor{blue}{62.5}\\
    SeqTrackv2-B384  &\textbf{\textcolor{cRed}{75.5}}&\textcolor{cGreen}{81.9}&\textbf{\textcolor{cRed}{91.8}} & &59.8&60.0&59.6\\
    SeqTrackv2-B256 &74.4&81.5&\textcolor{blue}{91.0} & &\textbf{\textcolor{cRed}{63.2}}&\textbf{\textcolor{cRed}{63.4}}&\textbf{\textcolor{cRed}{62.9}} \\
    \midrule[0.1pt]
    Un-Track~\cite{untrack} &72.1&\textcolor{blue}{82.0} &86.9 & &61.0&60.8&\textcolor{cGreen}{61.1}\\
    ViPT~\cite{vipt} &72.1&81.5 &\textcolor{cGreen}{87.1} & &59.4&59.6&59.2\\
    ProTrack~\cite{protrack} &65.1&80.1&80.2 & &57.8&57.3&58.3\\
    SPT~\cite{rgbd1k} &65.1&79.8&85.1 & &53.8&54.9&52.7\\
    SBT-RGBD~\cite{sbt} &70.8&80.9&86.4 & &-&-&-\\
    OSTrack~\cite{ostrack} &67.6&80.3&83.3 & &52.9&52.2&53.6\\
    DeT~\cite{depthtrack} &65.7&76.0&84.5 & &53.2&50.6&56.0\\
    DMTrack~\cite{vot2022} &65.8&75.8&85.1 & &-&-&-\\
    DDiMP~\cite{vot2020} &-&-&- & &48.5&56.9&50.3\\
    STARK-RGBD~\cite{Stark} &64.7&80.3&79.8 & &-&-&-\\
    KeepTrack~\cite{keeptrack} &60.6&75.3&79.7 & &-&-&-\\
    DRefine~\cite{vot2021} &59.2&77.5&76.0 & &-&-&-\\
    ATCAIS~\cite{vot2020} &55.9&76.1&73.9 & &47.6&45.5&50.0\\
    LTMU-B~\cite{LTMU} &-&-&- & &46.0&41.7&51.2\\
    GLGS-D~\cite{vot2020} &-&-&- & &45.3&36.9&58.4\\
    DAL~\cite{dal} &-&-&- & &42.9&36.9&51.2\\
    LTDSEd~\cite{VOT2019} &-&-&- & &40.5&38.2&43.0\\
    Siam-LTD~\cite{vot2020} &-&-&- & &37.6&34.2&41.8\\
    SiamM-Ds~\cite{VOT2019} &-&-&- & &33.6&26.4&46.3\\
    CA3DMS~\cite{ca3dms} &-&-&- & &22.3&22.8&21.8\\
    DiMP~\cite{DiMP} &54.3&70.3&73.1 & &-&-&-\\
    ATOM~\cite{ATOM} &50.5&59.8&68.8 & &-&-&-\\
    \bottomrule
    \end{tabular}
    }
  }
  \vspace{-3mm}
\end{table}

\textit{TNL2K.}
TNL2K~\cite{TNL2K} is a large-scale benchmark for vision-language tracking. TNL2K comprises 2,000 video sequences, with 1,300 for training and 700 for testing. Each video provides a language annotation along with bounding box annotation to indicate the target. In Sec.~\ref{subsec:sotav1}, we have evaluated our SeqTrack on TNL2K without utilizing the language modality. Here, we evaluate our multi-modal SeqTrackv2 on TNL2K using the language modality. As reported in Tab.~\ref{tab-sota-rgbl}, with the language modality, our SeqTrackv2 models get superior performance. All four SeqTrackv2 models perform better than previous vision-language trackers. SeqTrackv2-L256 obtains the best  AUC score of 62.7 \%, surpassing previous state-of-the-art tracker JointNLT by 5.8\%.

\textit{OTB99.}
OTB99~\cite{TNLS} is a small-scale benchmark for vision-language tracking.  As reported in Tab.~\ref{tab-sota-rgbl}, although the advantage is not be as significant as on other benchmarks, our method demonstrates competitive performance when compared with customized vision-language trackers. It is noteworthy that small-scale benchmarks are prone to overfitting, yet our models use same parameters and hyperparameters with other benchmarks.

\begin{table}[t]\normalsize
    \caption{State-of-the-art comparisons on RGB+Event tracking}
\label{tab-sota-rgbe}
\vspace{-2mm}
  \centering
\resizebox{1\linewidth}{!}{
  \setlength{\tabcolsep}{10mm}{
    \small
    \begin{tabular}{l|cc}
    \toprule
    \multirow{2}*{Method} & \multicolumn{2}{c}{VisEvent~\cite{visevent}}\\
        \cline{2-3}
 & AUC & P \\
    \midrule[0.5pt]
    SeqTrackv2-L384 &\textbf{\textcolor{cRed}{63.4}}&\textbf{\textcolor{cRed}{80.0}}\\
    SeqTrackv2-L256 &\textcolor{blue}{63.0}&\textcolor{blue}{79.4}\\
    SeqTrackv2-B384  &\textcolor{cGreen}{62.2}&\textcolor{cGreen}{79.3}\\
    SeqTrackv2-B256 &61.2&78.2 \\
    \midrule[0.1pt]
    ViPT~\cite{vipt} &59.2&75.8\\
    Un-Track~\cite{untrack} &58.9 &75.5\\
    OSTrack~\cite{ostrack} &53.4&69.5\\
    SiamRCNN\_E~\cite{SiamRCNN} &49.9&65.9 \\
    TransT\_E~\cite{transt} &47.4&65.0\\
    ProTrack~\cite{protrack} &47.1&63.2\\
    LTMU\_E~\cite{LTMU} &45.9&65.5 \\
    PrDiMP\_E~\cite{PrDiMP} &45.3&64.4 \\
    STARK\_E~\cite{Stark} &44.6&61.2 \\
    MDNet\_E~\cite{MDNet} &42.6&66.1 \\
    SiamCar\_E~\cite{Stark} &42.0&59.9 \\
    VITAL\_E~\cite{VITAL} &41.5&64.9 \\
    ATOM\_E~\cite{ATOM} &41.2&60.8 \\
    SiamBAN\_E~\cite{SiamBAN} &40.5&59.1 \\
    SiamMask\_E~\cite{SiamMask} &36.9&56.2 \\
    \bottomrule
    \end{tabular}
    }
  }
\end{table}

\begin{table}\footnotesize
    \caption{State-of-the-art comparisons on RGB+Language tracking}
\label{tab-sota-rgbl}
\vspace{-2mm}
  \centering
\resizebox{\linewidth}{!}{
  \setlength{\tabcolsep}{4.5mm}{  
  \small
  \begin{tabular}{l| cc c cc}
    \toprule
    \multirow{2}*{Method} & \multicolumn{2}{c}{TNL2K~\cite{TNL2K}} & & \multicolumn{2}{c}{OTB99}~\cite{TNLS}\\
    \cline{2-3}
    \cline{5-6}
    & AUC&P && AUC&P\\
    \midrule[0.5pt]
    SeqTrackv2-L384 &\textcolor{blue}{62.4}&\textbf{\textcolor{cRed}{66.1}} & &71.4&\textcolor{cGreen}{93.6}\\
    SeqTrackv2-L256 &\textbf{\textcolor{cRed}{62.7}}&\textcolor{blue}{66.0} & &70.3&91.0\\
    SeqTrackv2-B384  &\textcolor{cGreen}{59.4}&\textcolor{cGreen}{62.6} & &\textcolor{cGreen}{71.8}&93.4\\
    SeqTrackv2-B256 &57.5&59.7 & &71.2&\textcolor{blue}{93.9} \\
    \midrule[0.1pt]
JointNLT~\cite{JointNLT}	&56.9 &58.1 & &65.3&85.6 \\
DecoupleTNL\cite{DecoupleTNL}&56.7&56.0 & &\textcolor{blue}{73.8}&\textbf{\textcolor{cRed}{94.8}}\\
Zhao \emph{et al.}~\cite{zhao2023transformervision-languagetracking}	&56.0&- & &69.9&91.2 \\
VLT$_{TT}$~\cite{VLTTT}	&53.1&53.3 & &\textbf{\textcolor{cRed}{76.4}} &93.1 \\
CapsuleTNL~\cite{CapsuleTNL}	&- &- & &71.1&92.4\\
Li \emph{et al.}\cite{CTRTNL} &44.0&45.0 & &69.0&91.0\\
TNL2K-2~\cite{TNL2K}	&42.0&42.0 & &68.0&88.0 \\
SNLT~\cite{SNLT}	&27.6&41.9 & &66.6&80.4 \\
GTI~\cite{GTI} &-&- & &58.1&73.2\\
TransVG~\cite{transvg} &26.1&28.9 & &-&- \\
Feng \emph{et al.}~\cite{feng2019robust}	&25.0&27.0 & &67.0&73.0\\
RTTNLD~\cite{RTTNLD}	&25.0&27.0 & &61.0&79.0 \\
Wang \emph{et al.}~\cite{wang2018describe}	&-&-  & &65.8&89.1 \\
TNLS~\cite{TNLS}	&-&-  & &55.3&72.3 \\
OneStage-BERT~\cite{onestagebert} &19.8&-  & &24.6&32.2 \\
LBYL-BERT~\cite{onestagebert} &18.3&-  & &20.7&26.0\\
  \bottomrule
\end{tabular}
}}
\vspace{-3mm}
\end{table}

\subsection{Ablation and Analysis of SeqTrack. }
\label{subsec:ablation}

We use SeqTrack-B256 without the online template update mechanism as the baseline model in this ablation study. The result of the baseline is reported in Tab.~\ref{tab-ablation} (\#1).

\textit{Autoregressive v.s. Bidirectional.} Our method generates a sequence in an \emph{autoregressive} manner, which predicts the bounding box values one by one. We contrast this autoregressive method with another \emph{bidirectional} method that predicts all coordinate values simultaneously. In the bidirectional method, the input sequence comprises four special tokens akin to the \texttt{start} token. The decoder receives the sequence and predicts the four coordinates simultaneously in a batch. The causal attention mask is removed, enabling tokens to attend to each other bidirectionally. As demonstrated in Tab.~\ref{tab-ablation} (\#2), the bidirectional method significantly underperforms compared to the autoregressive one, highlighting the importance of maintaining the causal relationship between tokens for sequence modeling in tracking.

\begin{table}[t]
\definecolor{purple(x11)}{rgb}{0.63, 0.36, 0.94}
\definecolor{yellow(munsell)}{rgb}{1.0,0.988, 0.957}
\definecolor{green(colorwheel)(x11green)}{rgb}{0.0, 1.0, 0.0}
\definecolor{pink}{rgb}{1.0, 0.85, 0.85}
\centering
\caption{{Ablation Study on LaSOT~\cite{LaSOT} and GOT-10k~\cite{GOT10K}}. 
We use \textcolor{gray}{gray},  \textcolor{green!50}{green}, \textcolor{blue!50}{purple},  \textcolor{yellow!60}{yellow}, and \textcolor{pink}{pink} colors to denote baseline setting, framework, input to encoder, input to decoder, and tracking prior integration, respectively. $\Delta$ denotes the performance change (averaged over benchmarks) compared with the baseline.
}
\label{tab-ablation}
\vspace{-2mm}
\small
\resizebox{0.95\linewidth}{!}{
\setlength{\tabcolsep}{1mm}{
\begin{tabular}{l|c|cc|c}
\toprule
\# & Method &LaSOT &GOT-10k &$\Delta$\\
\midrule
\rowcolor{gray!15}
1 & Baseline &69.2 &73.7 &--\\
\rowcolor{green!5}
2 &Autoregressive $\rightarrow$ Bidirectional &64.8 &72.4 &\textbf{-2.9}\\  
\rowcolor{blue!5}
3 &Joint $\rightarrow$ Separate &62.0 &66.1 &\textbf{-7.4}\\
\rowcolor{yellow!5}
4 &$[x,$$y,$$w,$$h]$ $\rightarrow$ $[w,$$h,$$x,$$y]$ &67.1 &71.8 &\textbf{-2.0}\\  
\rowcolor{yellow!5}
5 &$[x,$$y,$$w,$$h]$ $\rightarrow$ $[x_{min},$$y_{min},$$x_{max},$$y_{max}]$ &68.3 &70.7 &\textbf{-2.0}\\ 
\ignore{
\rowcolor{yellow!5}
6 &Search $\rightarrow$ Search-Template &69.6 &73.3 &\textbf{-0.0}\\
\rowcolor{yellow!5}
7 &Search $\rightarrow$ Average &69.2 &72.2 &\textbf{-0.7}\\}
\rowcolor{yellow!5}
6 &Concat of Search-Template &69.6 &73.3 &\textbf{-0.0}\\
\rowcolor{yellow!5}
7 &Avg. of Search-Template &69.2 &72.2 &\textbf{-0.8}\\
\rowcolor{red!5}
8 &$+$ Likelihood-based Online Update &69.9 &76.1  &\textbf{+1.6}\\
\rowcolor{red!5}
9 &$+$ Naive Online Update &69.3 &73.1 &\textbf{-0.3}\\
\rowcolor{red!5}
10 &$-$ Window Penalty &68.8 &73.1 &\textbf{-0.5}\\
\bottomrule
\end{tabular}
}}
\vspace{-2mm}
\end{table}

\begin{figure}[t!]
\begin{center}
\includegraphics[width=1\linewidth]{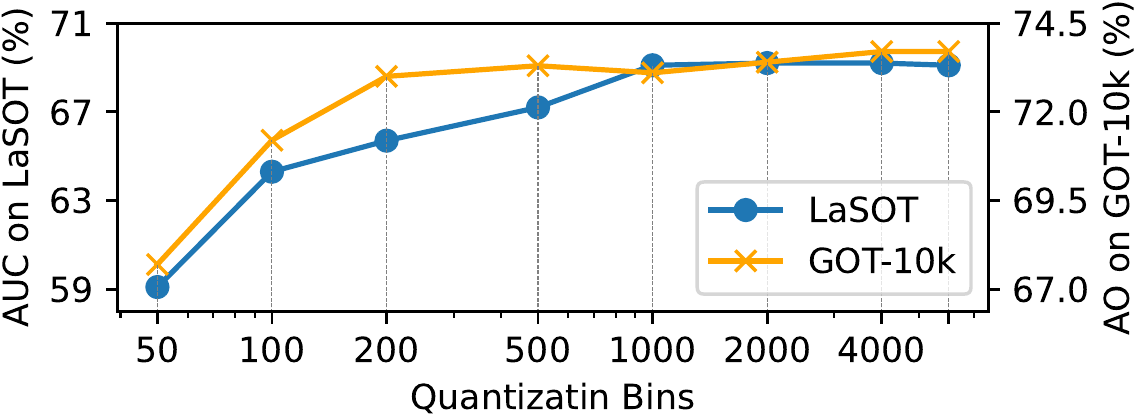}
\end{center}
    \vspace{-6mm}
   \caption{Influence of the number of quantization bins}
\label{fig:bins}
\end{figure}

\textit{Inputs of the Encoder.}
We evaluate various input strategies for the encoder. As outlined in Sec.~\ref{subsec:architecture}, the template and the search region are inputted into the encoder together. 
Subsequently, the encoder extracts their visual features in a \emph{joint} way. We contrast this with a \emph{separate} approach: akin to Siamese methods \cite{SiameseFC,SiameseRPN}, two weight-sharing encoders independently extract the features from the template and the search images, and then the extracted features are fed into the decoder. 
Tab.~\ref{tab-ablation} (\#3) shows that the separate feature extraction method lags behind the joint one by 7.2\% in AUC on the LaSOT benchmark. 
The underlying reason might be that the joint method enables the encoder to learn improved feature correspondence between the template and search images.

\begin{figure}[t!]
\begin{center}
\includegraphics[width=1\linewidth]{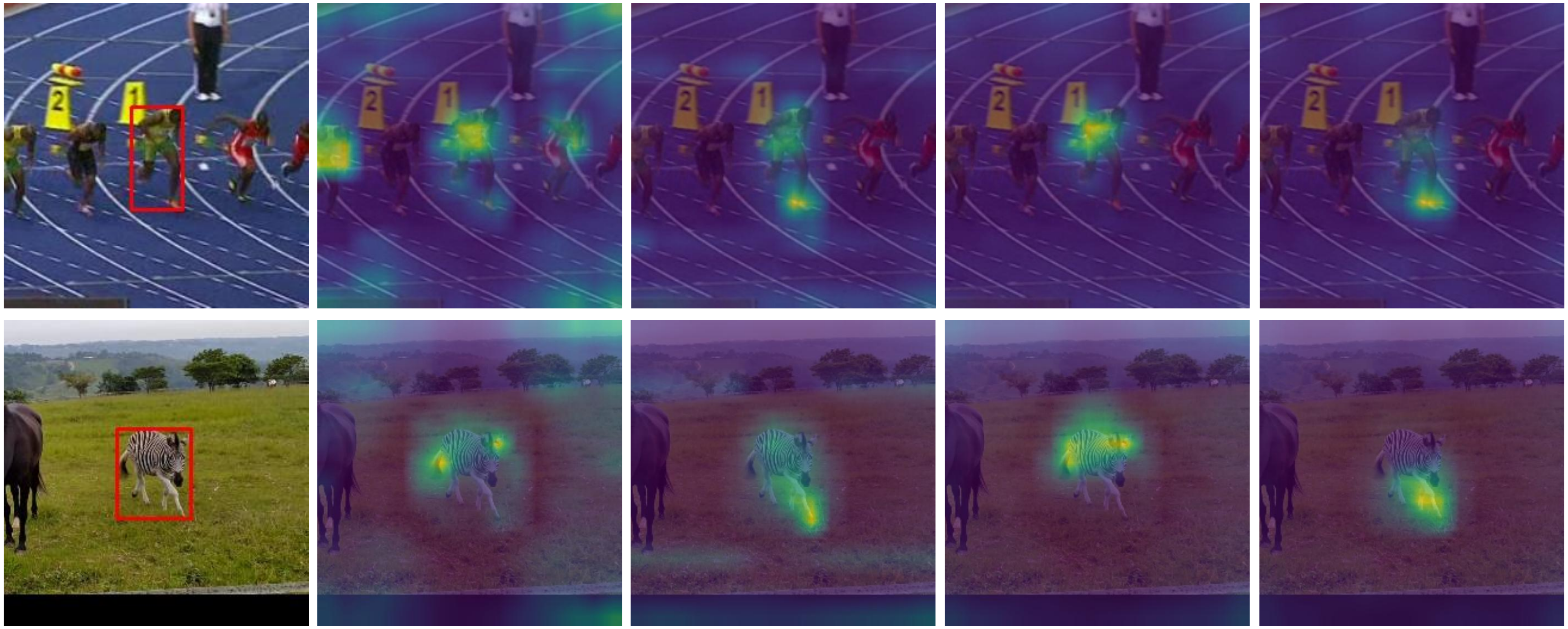}
\end{center}
    \vspace{-4mm}
   \caption{Decoder's cross attention to visual features when generating tokens. The first column is the search region image, and the second to last columns are the cross attention maps when $x,y,w,h$ are generated, respectively.}
\label{fig:attmap}
    \vspace{-3mm}
\end{figure}

\begin{table}[t]\normalsize
\centering
  \caption{Ablation study of the template size. The baseline setting is marked in \colorbox{gray!15}{gray}.}
  \label{tab-ablation_template}
      \vspace{-3mm}
\resizebox{1\linewidth}{!}{
  \setlength{\tabcolsep}{3mm}{
     \small
    \begin{tabular}{c|cc|cc}
    \toprule
    \#&Expansion Factor&Resolution&LaSOT~\cite{LaSOT}&GOT-10k~\cite{GOT10K}\\
    \midrule
    \rowcolor{gray!15}
    1 &4&$256$$\times$$256$&69.2&73.7 \\
    2 &2&$128$$\times$$128$&68.5&72.7 \\
    3 &2&$256$$\times$$256$&68.7&72.6 \\
    \bottomrule
    \end{tabular}
  }
}
  \vspace{-3mm}
\end{table}

\begin{table}[t]\normalsize
\centering
  \caption{Ablation study of the number of decoder blocks. Avg. denotes the performance averaged over benchmarks. The baseline setting is marked in \colorbox{gray!15}{gray}.}
  \label{tab-ablation_block}
    \vspace{-3mm}
\resizebox{1\linewidth}{!}{
  \setlength{\tabcolsep}{4mm}{
     \small
    \begin{tabular}{l|c>{\columncolor{gray!15}}ccccc}
    \toprule
    &1&2&3&4&5&6\\
    \midrule
    LaSOT~\cite{LaSOT}  &69.5&69.2&69.3&69.3&68.7&68.8\\
    GOT-10k~\cite{GOT10K}  &72.9&73.7&73.1&74.0&73.4&74.5\\
    Avg.  &71.2&71.5&71.2&71.7&71.1&71.7\\
    \bottomrule
    \end{tabular}
  }
}
  \vspace{-3mm}
\end{table}

\textit{Inputs of the Decoder.} 
We first compare different input sequences to the decoder. Tab.~\ref{tab-ablation} (\#4 and \#5) present two alternative formats of sequence:
1) $[$$w, $$h, $$x, $$y]$, where the model first generates object's scale and then its position coordinates;
and 2) $[x_{min}, $$y_{min}, $$x_{max}, $$y_{max}]$, where $[x_{min},$$y_{min}]$ denotes the top-left corner and $[x_{max},$$y_{max}]$ denotes the bottom-right one.
We observe that the default format $[x,$$y,$$w,$$h]$ yields the best result, as it aligns with human prior knowledge: first localizing object position and then estimating its scale.
We also explore the influence of the number of quantization bins $n_{bins}$, as shown in Fig.~\ref{fig:bins}. Increasing the number of bins $n_{bins}$ can improve the performance because the quantization error is reduced accordingly. The performance becomes saturated when $n_{bins}$ is larger than $4,000$, thus we set $n_{bins}$ to $4,000$.

Moreover, we compare different input visual features to the decoder. By default, we only feed the features of the search image into the decoder, as illustrated in Fig.~\ref{fig:framework}. Here, we compare it with two other alternatives: 1) the feature concatenation of the search and template images, and 2) the \emph{averaged} feature of the search and template images. For the first method, all the features are fed into the decoder. For the second method, the features are first projected into a 1D embedding and then fed into the decoder. From Tab.~\ref{tab-ablation} (\#6 and \#7), we observe that these two methods perform comparably to the default method that only uses search image features. Overall, the decoder does not appear to be sensitive to the form of input visual features.

\textit{Prior Knowledge Integration.} 
For the online update, our method utilizes the likelihood of generated tokens to select reliable templates.
As reported in Tab.~\ref{tab-ablation} (\#8), our method improves the tracking performance.
We also explore a naive online update method, where the dynamic template is updated without selection.
Tab.~\ref{tab-ablation} (\#9) demonstrates that this method obtains inferior performance. These results suggest that selecting reliable templates with our likelihood-based method is effective. 
For the window penalty, Tab.~\ref{tab-ablation} (\#10) demonstrates that the performance degrades without it.

\textit{Visualization of Cross Attention Map.}
To gain insights into how SeqTrack ``reads out" the target state, we visualize the cross attention (averaged over heads) of the last decoder block.
Fig.~\ref{fig:attmap} displays cross-attention maps as the model generates tokens. When generating the first token $x$, the attention appears relatively diverse.
As subsequent tokens are generated, attention quickly focuses on the target object.
Attention is more focused on key features, such as the person's arm and the zebra's tail when generating $x$ and $w$, and the foot when generating $y$ and $h$.

\textit{Template Size.}
As described in Sec. \ref{subsec:representation}, in prior trackers~\cite{SiameseFC,SiameseRPN,transt,Stark}, the size of template images is typically smaller than that of search images. In contrast, we use the same size for the template and search images. 
Here we compare different settings of the template size. 
Concretely, there are two steps to generate the template image: 1) expanding the target's bounding box by a certain factor to obtain a template region, and 2) resizing the template region to a certain resolution.
In our baseline setting, the expansion factor is $4$, and the resolution is $256$$\times$$256$.
The result of the baseline setting is shown in Tab.~\ref{tab-ablation_template} (\#1).
In Tab.~\ref{tab-ablation_template} (\#2), we scale down the baseline expansion factor and resolution, like prior trackers~\cite{SiameseFC,transt}, \emph{i.e.}, resolution of $128$$\times$$128$ and expansion factor of $2$. To exclude the effect of the resolution, we additionally compare a resolution-aligned setting with the expansion factor of $2$ and the resolution of $256$$\times$$256$, in Tab.~\ref{tab-ablation_template} (\#3).
Our baseline method gets the best result, demonstrating that adding more background information in the template is helpful for improving tracking performance.

\textit{Number of Decoder Blocks.}
We compare the number of transformer blocks in the decoder. Table~\ref{tab-ablation_block} shows that the model is not sensitive to this number. An underlying reason is that the information in the decoder is concise, thus a small capacity is sufficient.

\begin{table}[t]
\definecolor{purple(x11)}{rgb}{0.63, 0.36, 0.94}
\definecolor{yellow(munsell)}{rgb}{1.0,0.988, 0.957}
\definecolor{green(colorwheel)(x11green)}{rgb}{0.0, 1.0, 0.0}
\definecolor{pink}{rgb}{1.0, 0.85, 0.85}
\centering
\caption{Ablation Study of SeqTrackv2 on multi-modal benchmarks. 
We use \textcolor{gray}{gray},  \textcolor{green!50}{green}, \textcolor{blue!50}{purple},  \textcolor{yellow!60}{yellow}, and \textcolor{pink}{pink} colors to denote baseline setting, training tasks, instruction, training method, and prompting method, respectively. $\Delta$ denotes the performance change (averaged over benchmarks) compared with the baseline.}
\label{tab-ablationv2}
\vspace{-2mm}
\small
\resizebox{1\linewidth}{!}{
\setlength{\tabcolsep}{0.2mm}{
\begin{tabular}{l|c|cccc|c}
\toprule
\# & Method &LasHeR &VOT-RGBD22 &VisEvent &TNL2K &$\Delta$\\
\midrule
\rowcolor{gray!15}
1 & Baseline &55.8 &74.4 &61.2 &57.5 &--\\
\rowcolor{green!5}
2 & Multi-Task $\rightarrow$ Single-Task &57.2 &72.7  &61.2 &58.4 &\textbf{+0.2}\\  
\rowcolor{blue!5}
3 & $-$ Encoder Task Prompt &55.0 &74.8   &61.5 &57.3 &\textbf{-0.1}\\  
\rowcolor{blue!5}
4 & $-$ Decoder Task Prompt &54.8 &74.3 &61.1 &57.7 &\textbf{-0.3}\\  
\rowcolor{yellow!5}
5 & Add &46.2 &63.5  &52.4 &52.8 &\textbf{-8.5}\\
\rowcolor{yellow!5}
6 & Attention &50.6 &75.2  &60.6 &58.1 &\textbf{-1.1}\\
\rowcolor{yellow!5}
7 & Concat &43.4 &66.8  &53.0 &56.0 &\textbf{-7.3}\\
\rowcolor{red!5}
8 & Full Fine-Tuning &53.1 &71.4  &61.1 &60.6 &\textbf{-0.7}\\ 
\rowcolor{red!5}
9 & Equal $\rightarrow$  Proportional Videos  &55.4 &74.1  &61.1 &60.4 &\textbf{+0.5}\\
\rowcolor{red!5}
10 & Equal $\rightarrow$  Proportional Images  &53.2 &74.1  &59.3 &59.5 &\textbf{-0.7}\\
%\rowcolor{yellow!5}
%7 & Intra-Batch $\rightarrow$ Inter-Batch  & &  & & &\\

\bottomrule
\end{tabular}
}}
\vspace{-3mm}
\end{table}

\subsection{Ablation and Analysis of SeqTrackv2. }
\label{subsec:ablationv2}

We use SeqTrackv2-B256 as the baseline model in this ablation study. The result of the baseline model is reported in Tab.~\ref{tab-ablationv2} (\#1).

\textit{Multi-Task v.s. Single-Task.}
Our models are jointly trained on the four multi-modal tracking tasks, enabling them to address these tasks using a unified model and parameter set. We investigate the disparity between this unified model and individual single-task models. For the single-task models, we train an individual model for each multi-modal tracking task. As reported in Tab.~\ref{tab-ablationv2} (\#2), the single-task models yields only a marginal improvement of 0.2\% on average compared to the unified multi-task model. Overall, our unified model is able to perform on-par with individual single-task models. 

\textit{Task-prompt Token.}
We incorporate a task-prompt token to supply more task-centric information to the model, as described in Sec.~\ref{sec-taskprompt}. Here we explore the impact of the task-prompt token. 
In Tab.~\ref{tab-ablationv2} (\#3), we eliminate the task-prompt token from the encoder, leading to a slight performance dip. This could be attributed to the encoder's visual features already possessing substantial task information, reducing the need for additional supplementation. Despite this, the task-prompt token does contribute positively to performance. In Tab.~\ref{tab-ablationv2} (\#4), upon removing the decoder's task-prompt token, we observe a larger performance decline, underscoring the importance of introducing task-specific information during the decoding phase.
In summary, the task-prompt token enhances performance, with its need being greater in the decoding process than in encoding.

\begin{table}[t]\normalsize
\centering
  \caption{Ablation study of the low-rank dimension. Avg. denotes the performance averaged over benchmarks. The baseline setting is marked in \colorbox{gray!15}{gray}.}
  \label{tab-ablation_d}
    \vspace{-2mm}
\resizebox{1\linewidth}{!}{
  \setlength{\tabcolsep}{4mm}{
     \small
    \begin{tabular}{l|cc>{\columncolor{gray!15}}ccc}
    \toprule
    &8&16&32&64&128\\
    \midrule
    LasHeR~\cite{lasher}  &53.6&54.2&55.8&55.3&55.1\\
    VOT-RGBD22~\cite{vot2022} 
    &74.7&74.8&74.4&73.8&74.7\\
    VisEvent~\cite{visevent} &60.9&61.3&61.2&60.5&61.2\\
    TNL2K~\cite{TNL2K} &57.7&57.3&57.5&57.8&58.3\\
    Avg.  &61.7&61.9&62.2&61.9&62.3\\
    \bottomrule
    \end{tabular}
  }
}
  \vspace{-3mm}
\end{table}

\textit{Unified Interface.}
As detailed in Sec.~\ref{sec-interface}, we integrate the auxiliary-modal sequence into the visual features in a low-rank manner, that is,  by projecting both the auxiliary-modal and visual feature sequences from dimension $D$ to a lower dimension $d$. Here, we investigate the impact of the low-rank dimension $d$ setting. According to the results presented in Tab.~\ref{tab-ablation_d}, a too-low value of $d$ leads to diminished performance, but performance gains plateau once $d$ reaches 32. Thus, for considerations of efficiency and model size, we opted for $32$.

We have also investigated alternative choices for the unified interface. As shown in Table~\ref{tab-ablationv2} (\#5), we attempt to add the auxiliary-modal sequence directly into the encoder's visual features. In \#6, we explore the use of cross-attention modules to integrate the auxiliary-modal sequence. Entry \#7 documents our attempt to directly concatenate the auxiliary-modal sequence and the RGB-image patches sequence, and then input them into the encoder. Ultimately, we find that the default unified interface delivers superior performance.

\textit{Training Method.}
In Section~\ref{sec-train}, we detail our training methodology, wherein we keep the parameters of the base SeqTrack model frozen while exclusively training those of the newly-introduced unified interface alongside the task-prompt tokens. Additionally, we investigate the consequences of full fine-tuning. The results presented in Table~\ref{tab-ablationv2} (\#8) indicate that while full fine-tuning yields competitive performance, it remains inferior to our default training method. This discrepancy may be attributed to the fact that full fine-tuning disrupts the prior knowledge acquired from RGB tracking data.

In our default multi-task training strategy, we evenly mix the data from four multi-modal tasks. Furthermore, we explore allocating training data to different tasks based on the proportion of videos or images. The results corresponding to these strategies are shown in Tab.~\ref{tab-ablationv2} (\#9 and \#10). The results suggest that proportional strategies offer advantage for the task with a substantial amount of data (TNL2K, RGB+Language tracking). However, this advantage comes at the cost of the performance of tasks with smaller datasets. In contrast, the evenly-mixed strategy leads to more stable performance across different tasks.

\section{Conclusion}
\label{sec:conclusion}
This work proposes SeqTrack, a new sequence-to-sequence tracking framework for visual object tracking, and its advanced iteration, SeqTrackv2, which extends the framework for unified multi-modal tracking. SeqTrack uses a simple encoder-decoder transformer architecture, getting rid of complicated head networks and loss functions. SeqTrackv2 seamlessly integrates multi-modal tracking tasks into a unified model and parameter set. Through extensive experiments, we demonstrate the effectiveness of both SeqTrack and SeqTrackv2, achieving competitive performance compared to state-of-the-art trackers. We hope this work could catalyze more compelling research on sequence learning for visual tracking.

\emph{Limitation.}
One limitation of SeqTrack is that, despite achieving competitive performance, it is difficult to handle the cases when objects move out-of-view or are occluded by distractors, because the method does not have explicit re-detection modules.
Moreover, while SeqTrackv2 is constructed to unify four existing multi-modal tracking tasks, its effectiveness across new, untested modalities remains uncertain. A more promising avenue involves swiftly adapting the model to novel modalities with a few examples, a direction we intend to explore in future research.

\bibliographystyle{IEEEtran}
\bibliography{main}

\vfill

\end{document}